\documentclass{article}
\usepackage{hyperref}
\usepackage{url}
\usepackage[utf8]{inputenc} % allow utf-8 input
\usepackage[T1]{fontenc}    % use 8-bit T1 fonts
\usepackage{hyperref}       % hyperlinks
\usepackage{url}            % simple URL typesetting
\usepackage{booktabs}       % professional-quality tables
\usepackage{amsfonts}       % blackboard math symbols
\usepackage{nicefrac}       % compact symbols for 1/2, etc.
\usepackage{microtype}      % microtypography
\usepackage{amssymb}
\usepackage{xcolor,color}
\usepackage{framed}
\usepackage{booktabs}
\usepackage{bbm} 
\definecolor{shadecolor}{rgb}{1,0,0}
\usepackage{microtype}
\usepackage{optidef}
\usepackage{wrapfig}
\usepackage{multirow}
\usepackage{adjustbox}

\usepackage{amsthm}
\theoremstyle{plain}
\makeatletter
\newtheorem*{rep@theorem}{\rep@title}
\newcommand{\newreptheorem}[2]{%
\newenvironment{rep#1}[1]{%
 \def\rep@title{#2 \ref{##1}}%
 \begin{rep@theorem}}%
 {\end{rep@theorem}}}
\makeatother
\newtheorem{theorem}{\protect\theoremname}
\providecommand{\theoremname}{Theorem}
\newreptheorem{theorem}{Theorem}

\providecommand{\assname}{Assumption}
\newreptheorem{assumption}{Assumption}
\newtheorem{corollary}{\protect\corollaryname}
\providecommand{\corollaryname}{Corollary}
\newreptheorem{corollary}{Corollary}

\providecommand{\lemmaname}{Lemma}
\newreptheorem{lemma}{Lemma}

\providecommand{\remarkname}{Remark}
\newreptheorem{remark}{Remark}
\newtheorem{definition}{\protect\definitionname}
\providecommand{\definitionname}{Definition}
\newreptheorem{definition}{\definitionname}

\providecommand{\definitionname}{Conjecture}
\newreptheorem{conjecture}{\conjurnnam}

\providecommand{\definitionname}{Axiom}
\newreptheorem{axiom}{\axname}

\usepackage{microtype}
\usepackage{graphicx}
\usepackage{subfigure}
\usepackage{booktabs} % 

\usepackage{hyperref}

\usepackage[accepted]{icml2021}

\icmltitlerunning{Dynamic DP-SGD}

\begin{document}

\twocolumn[
\icmltitle{Dynamic Differential-Privacy Preserving SGD}



\icmlsetsymbol{equal}{*}

\begin{icmlauthorlist}
\icmlauthor{Jian Du}{equal,ant}
\icmlauthor{Song Li}{equal,sjtu}
\icmlauthor{Xiangyi Chen}{equal,um}
\icmlauthor{Siheng Chen}{sjtu}
\icmlauthor{Mingyi Hong}{um}
\end{icmlauthorlist}

\icmlaffiliation{ant}{Ant Group, Sunnyvale, CA}
\icmlaffiliation{sjtu}{Shanghai Jiao Tong University, China}
\icmlaffiliation{um}{University of Minnesota, MN}

\icmlcorrespondingauthor{Jian Du }{jd.jiandu@gmail.com}

\icmlkeywords{Machine Learning, ICML}

\vskip 0.3in
]



\printAffiliationsAndNotice{\icmlEqualContribution} % otherwise use the standard text.

\begin{abstract}
The vanilla Differentially-Private Stochastic Gradient Descent (DP-SGD), including DP-Adam and other variants, ensures the privacy of training data by uniformly distributing privacy costs across training steps. The equivalent privacy costs controlled by maintaining the same gradient clipping thresholds and noise powers in each step result in unstable updates and a lower model accuracy when compared to the non-DP counterpart.
In this paper, we propose the dynamic DP-SGD (along with dynamic DP-Adam, and others) to reduce the performance loss gap while maintaining privacy by dynamically adjusting clipping thresholds and noise powers while adhering to a total privacy budget constraint.
Extensive experiments on a variety of deep learning tasks, including image classification, natural language processing, and federated learning, demonstrate that the proposed dynamic DP-SGD algorithm stabilizes updates and, as a result, significantly improves model accuracy in the strong privacy protection region when compared to the vanilla DP-SGD. We also conduct theoretical analysis to better understand the privacy-utility trade-off with dynamic DP-SGD, as well as to  learn why Dynamic DP-SGD can outperform  vanilla DP-SGD.

% In this paper, we extend the Gaussian DP central limit theorem to calibrate the clipping value and the noise power for each individual step separately.
% We, therefore, are able to propose the dynamic DP-SGD, which has a lower privacy cost than the DP-SGD during updates until they achieve the same target privacy budget at a target number of updates.
% Dynamic DP-SGD, in particular, improves model accuracy without sacrificing privacy by gradually lowering both clipping value and noise power while adhering to a total privacy budget constraint. Extensive experiments on a variety of deep learning tasks, including image classification, natural language processing, and federated learning, show that the proposed dynamic DP-SGD algorithm stabilizes updates and, as a result, significantly improves model accuracy in the strong privacy protection region when compared to DP-SGD.
\end{abstract}

\section{Introduction}
% Deep learning has been driven by big data for the last decade without much concern about data privacy issues. 
% Data privacy protection is becoming increasingly important; not only are data breaches gaining public attention, but there are also data protection initiatives and data privacy laws in place, such as the General Data Protection Regulation (GDPR)  and the California Consumer Privacy Act (CCPA).
Publishing deep neural networks trained on private datasets poses a significant risk of data privacy leakage because the model embeds information about the training data. \citet{deep_leakage} and \citet{IDLG}, for example, both provide paradigms for breaching privacy and reconstructing training examples from published models. As a result, optimizers that protect the privacy of the model while training it are becoming increasingly important. 
Differential Privacy (DP)~\citep{DBLP:conf/icalp/Dwork06}, a gold-standard method for privacy-preserving computation that makes it nearly impossible for an adversary to single out a data record, is growing in importance and must be guaranteed to protect privacy.

%  Differential privacy (DP) is a provable and quantifiable method for privacy protection \citep{DBLP:conf/icalp/Dwork06}, which guarantees it nearly impossible for the adversary to differentiate  from two {\it neighboring data sets}.  
% Applying DP to maintain the privacy of the training data in deep learning is extremely difficult  due to the thousands to millions of repetitions\footnote{Around 300,000 steps is needed to train large models like GPT3.} of privacy  cost across the updates, resulting in a very high  noise power that completely degrades the learning process within a reasonable privacy preserving region.

By clipping the per-sample gradient and adding calibrated Gaussian noise in each step, Differentially-Private SGD (DP-SGD) \cite{DBLP:conf/ccs/AbadiCGMMT016} has been proposed to protect the model with the corresponding total privacy cost derived by composing the privacy costs of each step being less than a predefined privacy budget.
% In the MNIST dataset, for example, the moment-accountant based DP-SGD sacrifices accuracy by about $4.5\%$ with the privacy protection level $\epsilon=1.34$~\citep{bu2019deep}. Performance will continue to suffer as the level of privacy protection increases.
The clipping operation, which causes estimation bias to the true gradient, as well as the random noise, induce instability and even ramp-up of DP-SGD updates, as illustrated by the test shown in Fig.~\ref{fig:Gradient_Norm}. A high clipping threshold, intuitively, implies a small bias term, and vice versa. Under a fixed privacy, thereby, we can reduce bias by using large clipping thresholds, but at the costs of possible performance degradation due to increased noise power. 

In the vanilla DP-SGD, a constant clipping threshold is used, and choosing the proper clipping threshold, according to the literature, is more of an art than a science. 
This motivates us to study how to adapt to the dynamic of private gradient updates to adjust the clipping threshold and noise power so as to narrow the performance gap from both the practical implementation as well as theoretical understanding point of views.
In this paper, we make the following main contributions:

% To begin, motivated by the instability and even ramp-up of DP-SGD updates, as shown later in Fig.~\ref{fig:Gradient_Norm}, we investigate how to perform a tight DP accounting to enable proactive DP cost allocation for each  step to avoid unstable updates.
% Second, under this new privacy accounting framework, we propose instances of algorithms for proactive privacy budget allocation, such as the sensitivity-decay method, the growing-$\mu_t$ method, and a hybrid of the two known as dynamic DP-SGD.
% Third, we thoroughly validate the performance of dynamic DP-SGD for a variety of deep learning tasks, demonstrating  consistently significant accuracy improvements over existing methods while preserving privacy. The technical contributions are summarized below.

% In addition to novel algorithms, we investigate the central limit theorem of DP accounting for the dynamic DP-SGD to calibrate the noise power and clipping value. 

\begin{itemize}
% \item  In dynamic DP-SGD, allocating the  privacy budget to calibrate noise for each step is difficult due to the use of different mechanisms at each step.
% As a result, 
\item We extend the Gaussian DP's central limit theorem (CLT) 
to obtain an analytical composition results for non-uniform distributing privacy costs with the    Gaussian mechanisms.
The closed-form expression of the CLT result facilitates the dynamic clipping and noise  calibration along the updates under a predefined privacy budget constraint. We thereby propose the {\it sensitivity decay} and {\it growing-$\mu_t$} methods under this extended Gaussian DP CLT, and {show that combining these two methods leads to a novel {\it dynamic DP-SGD} algorithm with  additional performance gain in practice.}

% We investigate how to reduce the $\epsilon$ privacy cost compared to DP-SGD during updates until {\HC the target number of updates}, where they achieve the same target privacy budget under the recently proposed $\mu$-GDP criteria.

\item 
We perform a series of experiments and ablation studies on a variety of neural network tasks, including image classification, natural language processing, and federated learning, and  show that  the proposed dynamic DP-SGD effectively stabilizes gradient updates and consistently outperforms existing methods. With a strong privacy guarantee, i.e., at $\epsilon = 1.2 $, dynamic DP-SGD has a performance loss of only $2.49\%$ when compared to the non-DP version on MNIST dataset, and the loss is reduced to $1.72\%$ in the federated learning setting for a stronger privacy guarantee, i.e., $\epsilon=1$. 

\item 
We investigate analytically the impact of varying DP costs along the updates, and derive  the utility guarantee of the proposed DP-SGD method.
Mathematically, we demonstrate the utility  on both clipping thresholds and noise powers over iterations, and we gain insights into the
{ privacy-utility trade-off with dynamic DP-SGD.}
% impact factors {\color{red}[not clear what `impact factors' mean here]} for minimizing the model utility.  

% It demonstrates how to use the dynamic of DP-SGD to improve utility, despite the fact that much prior information, such as gradient dynamics and curvatures, are needed but may not be available in practice.
\end{itemize}

\section{Differential Privacy and Vanilla DP-SGD}
\label{sec:dp-sgd}

\begin{figure}[t]
    \centering
    \includegraphics[width = 0.33\textwidth]{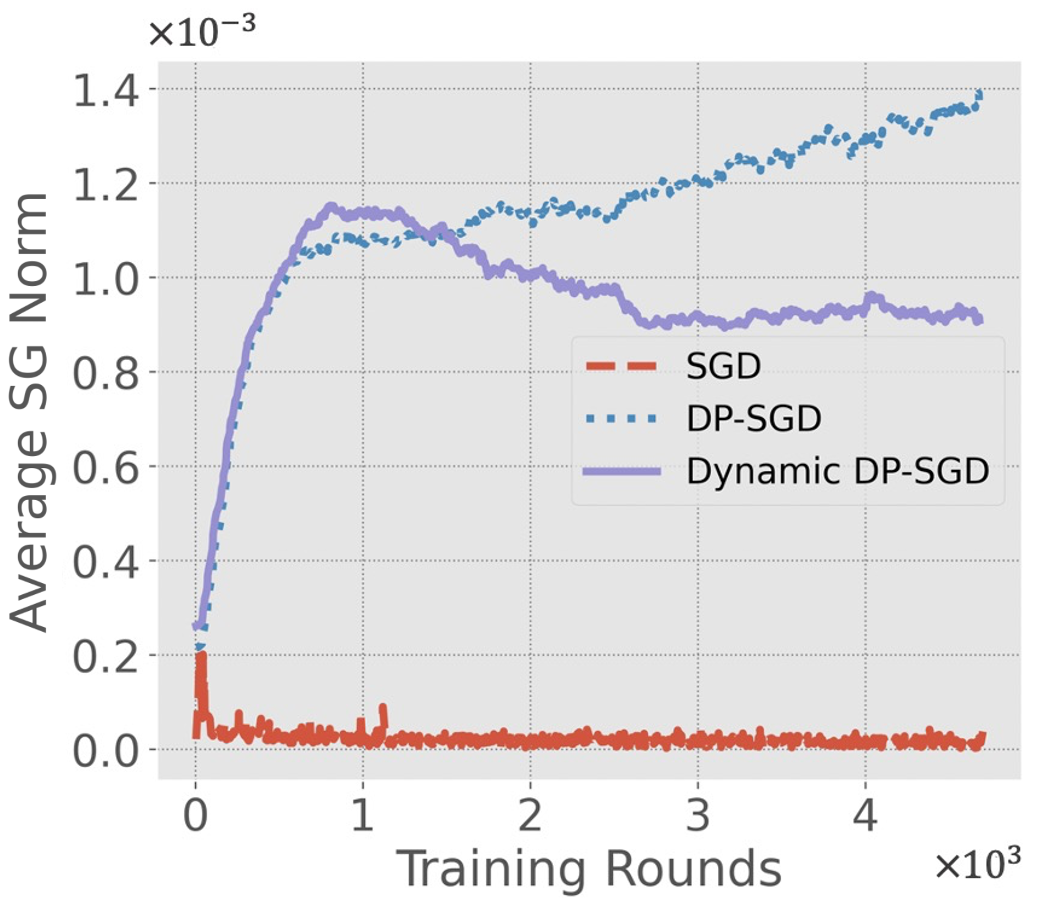}
    \caption{Experiments on MNIST with DP budget $(\epsilon,\delta)=(0.4,10^{-5})$ after $5\times 10^3$ steps for both DP-SGD and the proposed dynamic DP-SGD with the same corresponding  settings as that in Table~\ref{mnist_res}. In comparison to SGD, DP-SGD is unstable, and the stochastic gradient (SG) norm has a ramp-up period.
However, the SG norm is stabilized by the dynamic DP-SGD.
Additional experiments on the Fashion-MNIST is also provided in Appendix~\ref{app:unstable}. 
    }
    \label{fig:Gradient_Norm}
\end{figure}

Differential Privacy (DP) defines an upper bound of the privacy level by computing a pair of $(\epsilon, \delta(\epsilon))$.
A lower $\epsilon$ value leads to better privacy protection, but it potentially also makes the protected algorithm less useful.
The value $\delta$ can be interpreted as the probability of failing to achieve DP.
With the notion of the neighboring data sets, i.e., $X\sim X^{\prime}$, which differs by one {\it data record},
the formal $(\epsilon, \delta)$-DP definition is given below~\citep{DBLP:conf/icalp/Dwork06}.
\begin{definition} (($\epsilon, \delta(\epsilon)$)-DP Profile)
 A randomized algorithm $M(\cdot)$ gives $(\epsilon, \delta(\epsilon))$-differential privacy if for any pair of neighboring datasets $X\sim X^{\prime}$ and any event $E$ belongs to the range of $M$,
$$
\mathbb{P}(M(X) \in E) \leqslant \mathrm{e}^{\epsilon} \mathbb{P}\left(M\left(X^{\prime}\right) \in E\right)+\delta,
$$
\end{definition}
where the probability $\mathbb{P}(\cdot)$ is taken over the randomness of $M$, and $\epsilon
\geq 0$.
When $\delta = 0$, the algorithm is $\epsilon$-DP.
% The additive $\delta$ with $0\leq \delta\leq 1$  allows suppressing the long tails of the mechanism’s distribution where pure $\epsilon$-differential privacy guarantees may not hold. 
Intuitively, this means that we can't tell whether $M$ was run on $X$ or $X^{\prime}$ based on the results.
Thereby, an adversary will almost never infer the existence of any specific data record in the input data set.

In a Gaussian DP mechanism,  let $Y$ be the random variable following  Gaussian distribution with  $Y\sim \mathcal N\left(0, \sigma^2I_d\right)$ and $f: X^{n} \rightarrow \mathbb{R}^{d}$.
The Gaussian mechanism $ M(X)=f(X)+ Y$ follows the  DP profile~\citep{wang2019subsampled}:
\begin{equation}
\label{compute_mu}
\delta(\epsilon ; \mu)
% =1+G_{\mu}^{*}\left(-\mathrm{e}^{\epsilon}\right)
=\Phi\left(-\frac{\epsilon}{\mu}+\frac{\mu}{2}\right)-\mathrm{e}^{\epsilon} \Phi\left(-\frac{\epsilon}{\mu}-\frac{\mu}{2}\right),
\end{equation}
where 
\begin{equation}
\label{mu-def}
\mu = \frac{C}{\sigma},
\end{equation}
with  $C$ the sensitivity of $f(X)$ and $\Phi(\cdot)$  the Gaussian cumulative distribution function. 
The DP-SGD, which protects the model from revealing the training data privacy, is proposed in~\citet{DBLP:conf/ccs/AbadiCGMMT016} based on the Gaussian mechanism.
Let the model parameters be denoted by $\theta$.
and the loss function be $ L(\theta) = \frac{1}{N}\sum_{i=1}^N f(\theta, x_i)$ and the clipped stochastic gradient at step $t$ as $g_t = \frac{1}{p\left|X\right|}\big[
    \sum_{x\in X_{ t}}\text{CL}\left(g_x;C_t\right)
    \big]$ where $g_x \triangleq \frac{\partial f(\theta_t, x) }{\partial \theta_t} $.  
To calibrate the noise  required for DP, one can first clip the $\ell_2$-norm of the gradient by computing
\begin{equation}
\label{clipping}
  \widetilde g_x \triangleq 
\operatorname{CL}\left({g_x}; C\right) 
\triangleq 
g_x\cdot\min\left(1, \frac{C}{\left\|{g_x} \right\|}\right),
\end{equation}
and then add the  noise $\xi_t\sim \mathcal N(0,\sigma^2)$ with $\sigma^2 = \frac{C^2}{\mu^2}$. That is, the $t$-th step of the DP-SGD algorithm is given by:
\begin{align}
\label{dpsgd}
\text{DP-SGD:} \quad
\begin{split}
  \theta_{t+1} = \theta_{t} - \eta 
    \frac{1}{\left|X_{t}\right|} \left( 
    \sum_{x\in X_{ t}}\widetilde g_x
    + \xi_{t}\right).
\end{split}
\end{align}
Because of DP's post-processing property~\citep[Proposition~2.1]{dwork2014algorithmic}, protecting gradients provides the same level of privacy protection on the output model.
A closer examination of the DP-SGD process in (\ref{dpsgd}) reveals that the clipping-per-sample operator introduces biases to the original unbiased gradient estimate in the SGD updates if $C$ is not large enough. It is possible to have an unbiased gradient estimate by increasing $C$, if $C$ is greater than any $\ell_2$ norm of $\frac{\partial f_x}{\partial \theta}$. It does, however, result in an over-calibrated noise power.

\textbf{Motivations of this work}:
{In standard SGD updates for non-convex optimization, it has been shown that $\lim _{T \rightarrow \infty} \mathbb{E}\left[\left\| {\partial f(\theta_T, x) }/{\partial \theta_T}\right\|^{2}\right]=0$~\cite{bottou2018optimization}.}   {However, because the vanilla DP-SGD is followed by  {fixed} clipping threshold $C$ and constant noise power $\sigma^2$ as shown in Eq.(\ref{clipping}) and (\ref{dpsgd}), the ratio of noise power to the true gradient norm in 
Eq.~(\ref{dpsgd}) would continue to rise if $\lim _{T \rightarrow \infty} \mathbb{E}\left[\left\| {\partial f(\theta_T, x) }/{\partial \theta_T}\right\|^{2}\right]=0$, resulting in unstable updates.}
This hypothesis is supported by the experimental results in Fig.~\ref{fig:Gradient_Norm} as well as that displayed in Appendix~\ref{app:unstable}, which display the average coordinate stochastic gradient norm for all the data records in each step.
It is worth noting that the  stochastic gradient norm has a ramp-up period. Similar phenomenon  also appeared in the experiments in~\citep[Fig.~4]{Quantile}. The observation motivates us to investigate a dynamic DP-SGD to change  $C$ and $\sigma^2$ along with the  stochastic gradient updates in order to stabilize the iterations.
To facilitate the study,  in the following section, we  extend the Gaussian DP  CLT~\citep{dong2021gaussian} to compose the non-uniformly DP costs.

% \begin{figure}
%     \centering
%     \includegraphics[width = 0.5\textwidth]{figs/gradient_norm.jpg}
%     \caption{Dynamic DP-SGD can ensure a better convergence from a gradient norm perspective}
%     \label{fig:Gradient_Norm}
% \end{figure}

% \subsection{DP Accounting}
% Basic composition, simply sum the $(\epsilon,\delta)$,
% Strong composition,
% \citep{DBLP:conf/ccs/AbadiCGMMT016} proposes moment accountant with DP-SGD, by defining privacy cost function, moment accountant provides a relatively tight bound for privacy loss estimation.
% Rényi DP \citep{rdp}, use Rényi divergence to measure the distance between $f(X)$ and $f(X^{\prime})$.
% $\mu$-GDP measure the privacy cost from a hypothesis testing perspective. 
% Among these DP accounting framework, $\mu$-GDP provide the most tight estimation of privacy loss, thus making it require least noise power to achieve the same privacy level. In this paper, $\mu$-GDP is used as our baseline.
\section{Dynamic DP-SGD}
% The dynamic DP-SGD is presented in two parts: an extension of the CLT of GDP, specifically for dynamic noise and changing gradient clipping values, and the algorithms, which include growing-$\mu_t$, sensitivity decay, and dynamic DP-SGD.

% The DP parameter $(\epsilon, \delta)$ must be carefully determined, and existing theoretical analysis is frequently insufficient. In the following section, we work toward making DP machine learning practical by calibrating a reasonable noise power first providing refined privacy analyses and then designing a dynamic DP-SGD with a better convergence.
In the following, we evaluate the privacy cost using the Gaussian DP (GDP) framework~\citep{dong2021gaussian}, which measures the privacy profile $(\epsilon, \delta)$ in terms of $\mu = C/\sigma$ using Eq.~(\ref{compute_mu}) and (\ref{mu-def}). 
To make our paper self-contained, we include a preliminary of GDP  in Appendix~\ref{app:GDP-Pre}.
\subsection{Extended CLT for GDP}

Existing GDP CLT results~\citep{dong2021gaussian, bu2019deep} assume that each step satisfies the same $G_{\mu}$-DP, which is insufficient for the dynamic changing $C$ and $\sigma$ case. We investigate the extended GDP CLT to pave the way for  privacy accounting of dynamic DP-SGD, in which each step $t$ has different $C_t$ and $\sigma_t$, resulting in different $G_{\mu_t}$-DP costs to be composed.
Let $M_A$ and $M_B$ denote the compositions of $T$ steps updates of the training data sets $X$ and $X^{\prime}$, respectively, with $X\sim X^{\prime}$.
According to~(\ref{dpsgd}), each step consists of sub-sampling, denoted by $\text{PS}(\cdot)$ and local updates.
Then, we express  $M_A$ and $M_B$ by the composition of $T$-step updates, respectively,
$$ M_A \triangleq
M_{T} \circ \operatorname{PS}_T (X)\circ\cdots\circ M_{1} \circ \operatorname{PS}_1 (X)$$ $$ M_B \triangleq
M_{T} \circ \operatorname{PS}_T (X^{\prime})\circ\cdots\circ M_{1} \circ \operatorname{PS}_1 (X^{\prime}).$$
Consider the  sampling scheme $\operatorname{PS}(X)$ that each individual data sample $(x, y)$ is subsampled independently with probability $p$ from the training set to construct $X_t$.
It is shown \citep{bu2019deep} that a particular
 trade-off function $\text T$ as defined in Appendix~\ref{app:GDP-Pre} satisfies:
\begin{eqnarray}
\label{T-Tradeoff}
{\text T}(M_A, M_B)
\geqslant
\otimes_{t=1}^T
\left(p \cdot G_{\mu_t}+(1-p) \text { Id }\right),
\end{eqnarray}
where  $
G_{\mu_t}$  is a function of $\mathcal{N}(0,1)$ and $\mathcal{N}(\mu_t, 1)$  with $\mu_t = C_t/\sigma_t$
and  $\text { Id }(\alpha)=1-\alpha$.
The details of functions $\text T $
and $G_{\mu}(\alpha)$ are given in the appendix. 
The symbol $\otimes_{t=1}^T$ denotes the product of all the $T$ trade-off functions with the form $p \cdot G_{\mu_t}+(1-p) \text { Id }$, which is far from analytically computable.
 When all $\mu_t$ are the same, i.e., $\mu_t = \mu$,
\citet{bu2019deep} shows the CLT of the composition is given by 
\begin{equation}
\label{gdp-clt}
\mu_{\text{tot}} = p\sqrt{T(e^{\mu^2}-1)}.
\end{equation}
In the following corollary, we develop an extension of the above GDP CLT with the proof provided in Appendix~\ref{apped:theorem1}.

% Since we explore the potential of adjusting pravicy cost by adjusting $C_t$ and $\sigma_t$ in each round, existing central limit theorem in~(\ref{CLT1}) assuming the same $\mu_t$-GDP in each round cannot be applied directly.
% We obtain the CLT for the dynamic DP-SGD framework in the following Theorem with the proof provided in the appendix.
\begin{corollary}
\label{MU-CLT}
Consider a series of adaptive composition mechanisms $ M_t$ for $t\in [T]$, where
$ M_t$ is $ G_{\mu_t}$-DP, and each mechanism  works only on a  subsampled data sets by independent Bernoulli trial with probability $p$.  The trade-off function for $\lim _{T \rightarrow \infty} \otimes_{t=1}^T
\left(p\cdot  G_{\mu_t}+(1-p) \text { Id }\right)
$ in (\ref{T-Tradeoff}) approaches to $G_{\mu_{\text{tot}}}$-DP when $p\sqrt{T}$ is a constant, where   
\begin{equation}
\label{g2dp}
  \mu_{\text{\rm tot}}
  = p\cdot \sqrt{
  \sum_{t=1}^{T}
 \left(\mathrm{e}^{\mu_t^2}-1\right)},
\end{equation}
\end{corollary}
and $ \mu_{t} = C_t/\sigma_t$. Notably, when $\mu_t=\mu$ is substituted for $t\in [T]$, Eq.~(\ref{g2dp}) reduces to the CLT result in (\ref{gdp-clt}).
Similar to the original GDP CLT~\citep{dong2021gaussian, bu2019deep}, $\mu_{\text {\rm tot}}$ is an approximation of the privacy cost  after $T$ steps by CLT-type composition of Gaussian mechanism. We demonstrate numerically in Fig.~\ref{fig:rdp_gdp_comp} later in the experimental results section that only after hundreds of steps (much less than an epoch), this approximate result approaches its upper bound  computed by advanced 
RDP~\citep{balle2020privacy}. { For more details about RDP and the convergence verification of the GDP CLT, please refer to the explanation for Fig.~\ref{fig:rdp_gdp_comp}}.  

Corollary~\ref{MU-CLT} reflects that,
given the target privacy budget $(\epsilon,\delta)$ and the corresponding privacy parameter $ \mu_{\text{tot}}$  obtained by (\ref{compute_mu}),
we can proactively allocate privacy cost to each step $t\in [T]$ for a predefined total number of steps $T$, according to (\ref{g2dp}).  We can adjust both $C_t$ and $\sigma_t$ to control the privacy budget allocation.
Section~3.2-Section~3.4 detail the corresponding algorithms.

\subsection{Growing-$\mu_t$ Method}
{ According to the  discussion at the end of Section~\ref{sec:dp-sgd}, it is expected that the stochastic gradient norm decreases in expectation} along the updates. However, vanilla DP-SGD with constant $\sigma$ leads to an increasing noise power to true gradient ratio with proper $C$, under which  the stochastic gradient would not be clipped after some $t$ if its norm becomes smaller than $C$ in the ideal case. If so, it is natural to reduce the noise power to stabilize the gradient updates. { Thereby, we investigate  a method to reduce the noise power $\sigma_t$ while keeping $C_t=C$. According to Eq.~(\ref{mu-def}), $\mu_{t+1}\geq \mu_t$ for all $t\in [T]$ in this case, and we therefore term this method as Growing-$\mu_t$}.  
With (\ref{g2dp}) in mind, we  control the privacy cost rate by adjusting a hyper-parameter $\rho_{\mu}$ with 
\begin{equation}
\label{c_decay}
     \rho_{\mu}\triangleq\frac{\mu_{T}}{\mu_{0}},
    \quad \rho_\mu\geq 1.
\end{equation}
Then the dynamic $\mu_t$ is given by
\begin{equation}
\label{mut}
\mu_t = (\rho_{\mu})^{t/T}\cdot \mu_0, \quad \forall t\in [T].
\end{equation}

Given the total privacy budget $(\epsilon, \delta)$, the equivalent  privacy parameter   $\mu_{\text{tot}}$ is obtained according to (\ref{compute_mu}).
Then the remaining problem is to determine the initial state $\mu_0$. Once  $\mu_0$ is obtained, the entire $\{\mu_t\}$ sequence can be generated according to (\ref{mut}).
By substituting (\ref{mut}) into  (\ref{g2dp}), we obtain 
\begin{equation}
\label{mu0}
  \mu^2_{\text{tot}}
  = p^2\cdot  
  \sum_{t=1}^{T}
\left(\mathrm{exp}\left\{\left((\rho_{\mu})^{t/T}\cdot \mu_0 \right)^2\right\}-1\right).
%  \left((\rho_{\mu})^{t/T}\cdot \mu_0 \right).
\end{equation}

Because the above equation is transcendental when $\rho_{\mu}>1$, there is no closed-form solution for $\mu_0$. Nonetheless, the r.h.s of (\ref{mu0}) is monotone increasing w.r.t. $\mu_0$, and we can thus solve it efficiently using a numerical method such as binary search.
The computation of $\mu_t$ is summarized in Algorithm~\ref{alg:mu_allocation}\footnote{The  privacy cost rate form in (\ref{c_decay}) is not the only one; Trying different forms requires more heuristic work, and it also makes the theoretical analysis of privacy-utility trade-off later more involved. }. The noise power at iteration $t$ can be calculated using a specific expression of $\mu_0$ and $\rho_\mu$:
\begin{equation}
    \sigma_t = \frac{C}{\mu_0}(\rho_\mu)^{-\frac{t}{T}}, \quad \rho_{\mu}>1,\;{ \forall~t\in[T]}.
    \label{noise_power_decay}
\end{equation}

\begin{figure}[t]
    \centering
    \includegraphics[width = 0.3\textwidth]{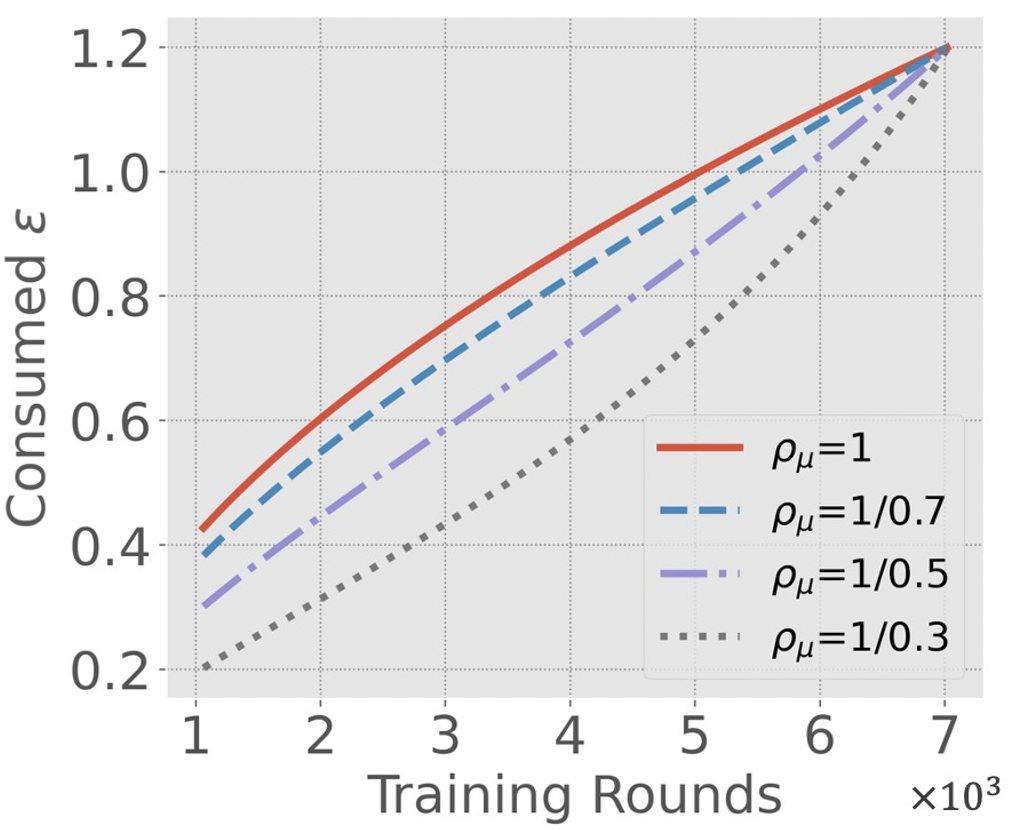}
    \caption{Consumption of privacy budget with different $\mu_t$ increasing rates; $\epsilon =1.2$.  } 
    \label{fig:eps_consumption}
\end{figure}

One numerical example is shown in Fig.~\ref{fig:eps_consumption}. Given the total DP budget  $\epsilon=1.2$, growing-$\mu_t$ gives the freedom to adjust the privacy cost rate,  which is the slope of the curve. 
In Fig~\ref{fig:eps_consumption}, we demonstrate the privacy budget consumption curve for different $\rho_\mu$. 
% We draw the figure starting from iteration 1000 as $\mu$-GDP requires CLT, which is more precise with large composition times. 
The solid line ($\rho_\mu = 1$) represents vanilla DP-SGD with evenly distributed noise power along with the step updates. With growing-$\mu_t$, we can now realize any $\mu$ consumption process under the constraint of the total DP budget as shown by the dashed lines. Specifically, the growing-$\mu_t$ slows consumption in the early rounds and accelerates consumption in the later rounds.

% \begin{wrapfigure}{R}{0.4\textwidth}
%     \begin{minipage}{0.4\textwidth}
%     \vspace{-25pt}
\begin{algorithm}[t]

\caption{$\mu_t$ computation }
\label{alg:mu_allocation}
\begin{algorithmic}[1]
\REQUIRE    privacy budget $ (\epsilon, \delta)$,  $T$, $\rho_{\mu}$.
\STATE Compute $ \mu_{\text{tot}}$ corresponding to $(\epsilon, \delta)$ according to (\ref{compute_mu}).
\STATE Compute $\mu_0$ in  (\ref{mu0})   by binary search.
% \STATE $\rho_\mu = \rho_{\mu}^{\frac{1}{T}}$  with  $\rho_\mu>1$
\STATE Compute $\sigma_t $ according to (\ref{noise_power_decay}) for all $t\in [T]$.
% \STATE Compute $\sigma_t = \frac{C}{\mu_0}\rho_\mu^{-\frac{t}{T}} $
\end{algorithmic}
\end{algorithm}
% \vspace{-50pt}
% \end{minipage}
% \end{wrapfigure}
% \begin{wrapfigure}{R}{0.4\textwidth}
%     \begin{minipage}{0.4\textwidth}
%     \vspace{-20pt}
% \vspace{-50pt}
% \end{minipage}
% \end{wrapfigure}

% With the total privacy budget constraint, we begin by increasing the DP budget cost rate in terms of increasing $\mu_t$ by decreasing the Gaussian noise power with the training steps. 
% In general, the sequence of $\mu_t$ can be increased in any pattern. In this paper, we set the  sequence  increasing in a linear manner with a hyper-parameter $\rho_\mu>1$ such that
% \begin{equation}
%      \rho_\mu=\frac{\mu_{T}}{\mu_{0}}.
%     \label{mu_grow}
% \end{equation}
% with the dynamic update
% \begin{equation}
% \label{mut}
% \mu_t = (\rho_{\mu})^{t/T}\cdot \mu_0, \quad \forall t\in [T].
% \end{equation}

\vspace{10pt}
\subsection{Sensitivity-Decay Method}
{ When the required level of privacy protection is high, i.e., a small $\mu_{\text{tot}}$, it requires a large $1/\mu_t =\sigma_t/ C_t$ according to Eq.~(\ref{g2dp}). If $ C_t$ is larger than the (stochastic) gradient norm, the noise, which is calibrated by $C_t$, will dominate the true gradient and could lead to unstable DP-SGD updates as shown in Fig.~\ref{fig:Gradient_Norm}.
Based on this observation and considering  the previous discussion that the stochastic gradient norm should be decreased in expectation to ensure the convergence, we design to reduce  $C_t$ across the updates.  Compare to the constant $C$ case in vanilla DP-SGD, it will reduce the required noise power after a particular $t$ when $C_t<C$.}

% \begin{wrapfigure}{R}{0.45\textwidth}
%     \begin{minipage}{0.45\textwidth}
% \begin{algorithm}[H]
% \caption{Sensitivity Decay }
% \label{alg:sens_decay}
% \begin{algorithmic}[1]
% \REQUIRE  Training iterations $T$,Initial sensitivity $C_1$, Target total sensitivity decay ratio $\rho_{c}$,
% %\STATE $ \rho_{c, T} = \frac{C_T}{C_{1}} $  with  $\rho_{c, T}<1$
% \STATE Sensitivity decay rate $\rho_c = \rho_{c}^{\frac{1}{T}}$
% \STATE $C_t = \rho_c^{t}\cdot C_0$
% \end{algorithmic}
% \end{algorithm}
% \end{minipage}
% \end{wrapfigure}
% In DP-SGD, the sensitivity of gradient is described by the $\ell_2$ norm. To determine required quantity of noise in each iteration, we need to bound of sensitivity by a clipping value $C$. In practice, in the training course of neural network, the sensitivity of gradients generated by one sample is clamped by the clipping value $C$. Previous works apply the same clipping value during the whole training process or introduce dynamic clipping based on statistical information like quantile. The same clipping value will result in the divergence of gradient norm which will be explained later. Quantile based clipping requires additional DP mechanism on quantile number. In this paper, based on observations on the behavior of gradient norm, we propose to use dynamic sensitivity for different training iterations. Specifically, we propose to apply a decay function on sensitivity as training goes on. 
Towards this end, we 
set the sequence of clipping values as follows
\begin{equation}
    % \frac{C_t}{C_{t-1}} = \rho_c \quad \text{with} \quad \rho_c<1
    C_t =   (\rho_{c})^{-\frac{t}{T}}\cdot C_0, \quad \rho_c\geq 1, \; { \forall~t\in[T]}.
    \label{sens_decay}
\end{equation}
Different from the growing-$\mu_t$ method, we fix $\mu_t=\mu_0$. By further substituting $\mu_t=\mu_0$ into (\ref{g2dp}) in Theorem~1,  we have the closed-form solution:
\begin{equation}
    \mu_0 = \sqrt{\log\left(\frac{\mu_{\text{tot}}^2}{p^2T}+1\right)}, \quad t\in[T].
    \label{mut_compute}
\end{equation}
With (\ref{sens_decay}) and (\ref{mut_compute}), we can calibrate the noise power at each step by:
\begin{equation}
     \sigma_t = \frac{C_0}{\mu_0}(\rho_c)^{-\frac{t}{T}}, \quad \rho_c>1, \quad t\in[T].
    \label{sigmat_compute_sens_decay}
\end{equation}

\subsection{Dynamic DP-SGD}
% \begin{wrapfigure}{R}{0.5\textwidth}
%     \begin{minipage}{0.5\textwidth}
%     \vspace{-30pt}
\begin{algorithm}[b]
%\begin{algorithm}[htbp]
\caption{Dynamic DP-SGD Algorithm}
\label{alg:NADP}
\begin{algorithmic}[1]
\REQUIRE  DP budget $(\epsilon,\delta)$, sampling rate $p$ and hyper-parameters:  $\rho_{\mu}$, $\rho_{c}$ and $C_0$.
% \STATE Compute $\mu_{\text{tot}}$ according to (\ref{compute_mu}) \Note{SC: do we need this step?}
\STATE Compute 
$\mu_0 $ in Algorithm~\ref{alg:mu_allocation} 
\FOR{$t=1, \ldots, T$}
%\STATE Compute $C_t$,$\mu_t$ according to \ref{alg:dynamic_noise}
% \STATE Compute $C_t =   (\rho_{\mu_2})^{-\frac{t}{T}}\cdot C_0$.
\STATE Compute $C_t =   (\rho_{c})^{-\frac{t}{T}}\cdot C_0$ in (\ref{sens_decay})
\STATE Calibrate noise :
$\sigma_t = \frac{C_0}{\mu_0}(\rho_{\mu}\cdot\rho_c)^{-\frac{t}{T}}$
% according to 
% (\ref{sigmat_compute_sens_decay}) by replacing  $\rho_c$ with $\rho_{\mu}\cdot\rho_c$. %by substituting $C_t$ into (\ref{noise_power_decay})
% \STATE $\sigma_t = \frac{\rho_{\mu_2}}{\rho_{\mu_2}}\frac{C_0}{\mu_0}$
\STATE 
Sample $ X_t\in X$ with sampling rate $p$ and sample noise $\xi_{t}\sim \mathcal N(0,\sigma^2_{t}I)$. 
\STATE Compute:
% \STATE
%  $\theta_t^{(\ell)} =  \theta_{t-1}^{(\ell)} - \eta 
%     \frac{1}{\left|X_t\right|} \left(\widetilde g_t + \xi_{t}\right)$ with 
%     $\xi_{i,t}\sim \mathcal N(0,\sigma^2_{i,t})$
    $\theta_{t+1} =  \theta_{t} - 
    \frac{\eta }{\left|X_t\right|}\big[\xi_{t}+$
    $ 
    \sum_{x\in X_{ t}}\text{CL}\left(g_x;C_t\right)
    \big]$ with $\text{CL}(\cdot)$ given in (\ref{clipping}) 
\ENDFOR
\end{algorithmic}
\end{algorithm}
% \vspace{-50pt}
% \end{minipage}
% \end{wrapfigure}

Because the noise calibration is based on $C_t$, we incorporate the growing-$\mu_t$ method into the sensitivity-decay method and refer to this new one as dynamic DP-SGD.
It maintains the same $\mu_t$ increasing rate as the previous growing-$\mu_t$ method while having a faster noise decay rate than the sensitivity-decay method.
We summarize the dynamic DP-SGD algorithm in Algorithm~\ref{alg:NADP} and conduct extensive experiments to show how dynamic DP-SGD improves performance.

\textbf{Remark on the dynamic private version of other optimizers:}
DP version of  other optimizers besides SGD can benefit from the proposed dynamically changing clipping thresholds and noise powers. 
A private version of Adam, which is refer to as  dynamic DP-Adam, for example, can be obtained similar by   first computing the average per-sample gradient using Line~6 in Algorithm~\ref{alg:NADP}, i.e., $\frac{1 }{\left|X_t\right|}\big[\xi_{t}+ 
    \sum_{x\in X_{ t}}\text{CL}\left(g_x;C_t\right)
    \big]$. According to the post-processing property of DP~\citep[Proposition~2.1]{dwork2014algorithmic}, the first and second moments of the gradient can be computed based on this DP guaranteed result, and accordingly the DP version of Adam can be computed with details provided in Appendix~\ref{app:da-adam}. Similarly, we can derive DP version of AdaGrad, RMSProp, etc.

% \begin{wrapfigure}{R}{0.45\textwidth}
%     \begin{minipage}{0.45\textwidth}
% \begin{algorithm}[H]
% \caption{Dynamic DP }
% \label{alg:dynamic_noise}
% \begin{algorithmic}[1]
% \REQUIRE  Training iterations $T$, Target $\mu$, Total $\mu_t$ increasing ratio $\rho_{\mu}$, Initial sensitivity $C_0$, Total sensitivity decay ratio $\rho_{c}$
% \STATE Binary Search to compute $\mu_0$ according to (\ref{g2dp}) and (\ref{mu_grow})
% % \STATE $\rho_\mu = \rho_{\mu, T}^{\frac{1}{T-1}}$, $\rho_c = \rho_{c}^{\frac{1}{T}}$
% \STATE $\mu_t = \mu_0 \cdot \rho_\mu^{\frac{t}{T}} $,$C_t = C_1\cdot\rho_c^{\frac{t}{T}}$
% \STATE $\sigma_t = \frac{C_t}{\mu_t} $
% % \STATE $\sigma_t = C_t/\mu_t$
% % \STATE $\sigma_t = \sigma_1 \cdot (\rho_c/ \rho_\mu)^{t-1} $
% \end{algorithmic}
% \end{algorithm}
% \end{minipage}
% \end{wrapfigure}

\section{Experiments}
\label{sec:experiments}
\subsection{Datasets, Models and Benchmarks}
\textbf{Datasets}: To conduct a comprehensive test of the dynamic DP-SGD performance, we run experiments with neural network models
% \footnote{We do not test the results on large models such as BERT due to the high memory over-head of the per-sample clipping operation in DP-SGD, which is beyond the scope of this paper. Please see \citet{li2021large} for information on how to improve the memory efficiency and throughput of the DP-SGD.} 
including MLP, CNN, LSTM, Federated Learning, and VGG-11
on the following 6 datasets: MNIST, FashionMNIST, IMDB, NAME,  InfiniteMNIST, and chest radiographs
from the Paediatric Pneumonia dataset~\citep{kermany2018identifying}. In  Appendix~\ref{apped:dataset},  we describe each data set, the corresponding neural network model, as well as the parameter settings for each experiment. We  also provide details about the dynamic DP-SGD federated learning algorithm for the corresponding federated learning experiments. 

\textbf{Benchmarks}:  We compare our proposed dynamic DP-SGD, growing-$\mu_t$ method, and sensitivity-decay method to the four benchmarks listed below.
 
\noindent{(i) \it{SGD without DP}}: The SGD method serve as the upper bound of model  accuracy in the absence of clipping and additive noise. 
% Please note that  the DP-SGD and proposed dynamic DP-SGD can also be used to obtain the DP-Adam counterparts of an Adam optimizer due to DP's postprocessing properties~\citep{dwork2009differential}. 
For the IMDB data set, we apply the Adam to compute the performance upper bound.

\noindent{(ii) \it{Vanilla DP-SGD }}: Under the same privacy budget constraint, Vannilla DP-SGD could require different amount of noise due to different DP accounting methods.  It is shown by~\cite{gopi2021numerical} that the GDP CLT provides a tighter composition bound for DP-SGD and even 
underestimate the privacy cost compared to other methods including the  moment accountant method~\citep{DBLP:conf/ccs/AbadiCGMMT016}. As a result, the calibrated noise power is lower than that obtained by moment accountant, resulting in a higher accuracy. Thereby, it provides the best performance that the vanilla DP-SGD can achieve with a predefined privacy budget. We therefore have vanilla DP-SGD under the privacy accounting by GDP CLT  to sever as a baseline.

\noindent{(iii) {{\it $\rho$-zCDP-SGD}} ~\citep{yu2019differentially}}: This paper is relevant to our work as it proposes decaying the noise power during the SGD and computing the privacy loss using the $\rho$-zCDP. However, only parallel composition is considered, with no regard for DP amplification by subsampling.

% In detail, for $\rho$-zCDP, given a sequence of mechanism ($\rho_1,\rho_2...\rho_n$) resulting on the sequence of randomized partitions of the training data, the composed mechanism within one training epoch satisfies $\mathop{max}\limits_{i}$ $\rho_i$-zCDP.
% Denote two composed mechanisms among different epochs: A1 and A2, satisfying $\rho_1$-zCDP and $\rho_2$-zCDP respectively, their sequential composition A = (A1, A2) satisfies ($\rho_1+\rho_2$)-zCDP. Given $\delta$ and $\rho$ in $\rho$-zCDP, then it provides ($\rho + \sqrt{2\rho\log(1/\delta), \delta)}$)-DP. With these three properties, like in dynamic DP, we can now compute initial $\sigma$ with binary search.

\noindent{(iv) \it{tCDP-SGD }~\citep{CDP}}:
This more recently work proposed decaying the noise power for DP-SGD training as well as analyzing the DP composition and subsampling amplification under the truncated concentrated differential privacy (tCDP) framework.

\textbf{Parameters}: 
We concentrate on the strong privacy guarantee, with the smallest $\epsilon$ set to be 0.1. We set different $\epsilon$'s denoted in each table due to different training data sizes and learning problem hardness; and $\delta = 1/(10|X|)$, where $|X|$ is the training data sample size. It is worth noting that we observed a significant performance degradation of DP-SGD when performing IMDB tasks. As a result, we also test a relatively large range of  $\epsilon$, i.e. $\epsilon\in [0.5,9]$. 
The hyperparameters $\rho_c$ and $\rho_{\mu}$ are swept in the following predefined sets:
$
    1/\rho_{c}, 1/\rho_{\mu} \in  \{0.1,0.2,0.3,0.4,0.5,0.6,0.7,0.8\}.
    % \\
    % \centering
    % 1/\rho_{\mu} \in \{0.3,0.4,0.5,0.6,0.7,0.8\}.
$
Other hyper-parameters are detailed in Appendix~\ref{apped:dataset}\footnote{The code will be available shortly at github.com/dynamic-dp.}.

\subsection{Results Analysis}
Experiment results for the above six different datasets on MLP, CNN, LSTM,  Federated Learning, and VGG models are shown in Table~1-Table~5, along with the performance of benchmarks and proposed methods.
In general, the proposed extended GDP CLT supports the privacy accounting for dynamic clipping and noise power decay in dyamic DP-SGD. 
Separate experiments are carried out with the proposed growing-$\mu_t$ method, sensitivity-decay method, and dynamic DP-SGD method. The results in Table~1-Table~5 show consistently that all the three proposed methods improve performance when compared to the static noise GDP method~\cite{bu2019deep}.
In particular, 
$\mu_t$ grows at the expense of early convergence speed in order to achieve higher accuracy, while sensitivity decay ensures more stable convergence. This explains why the sensitivity decay outperforms that of growing-$\mu_t$. The dynamic DP-SGD, a combination of the two, improves performance while causing no additional privacy loss, as demonstrated by the Dynamic DP results in each table.

Specifically, experiments with different privacy budgets were conducted, and the results show that the stronger the privacy protection required (i.e., the lower $\epsilon$ value), the better the proposed dynamic DP-SGD method performs. For example, when $\epsilon = 0.4$ for the MNIST, \cite{yu2019differentially} and \cite{zhang2021adaptive} adopt noise decay, but their model fails to learn due to large calibrated noise power by the loose DP compositions. Our method, on the other hand, outperforms the vanilla DP-SGD method by a large margin, achieving $3.17\%$ when $\epsilon = 0.4$ and even $4.6 \%$ for the LSTM network on NAME ($\epsilon=1$) shown in Table~3.

It is worth noting that if the DP accountant is not tight enough, the noise power required will be overestimated, resulting in the noise dominating the gradient and causing the network to fail to learn. For example, the $\rho$-zCDP-SGD method  with loose DP compositions performs significantly worse in the high privacy guarantee region due to overestimated noise.  As a result,  we only replicate its CNN model results on the MNIST and FashionMNIST datasets in Table~1, but omitting the meaningless results on other datasets.
The extended GDP CLT , on the other hand, provides a detailed examination of DP amplification through subsampling as well as DP composition. As a result, for the same privacy budget, more precise noise power is calibrated. 
% Specifically, even when there are no dynamics for DP-SGD updates, it outperforms the noise decay method's accuracy denoted as $\rho$-zCDP-SGD and tCDP-SGD. {\red[I felt that the discussion in the last couple of sentences is a bit out of the place (i.e., no numerical results to show this, and we don't need to show this, right?)]}

\subsection{Hyper-parameter sensitivity}
\begin{figure}[htb!]
    \centering
    \includegraphics[width = 0.45\textwidth]{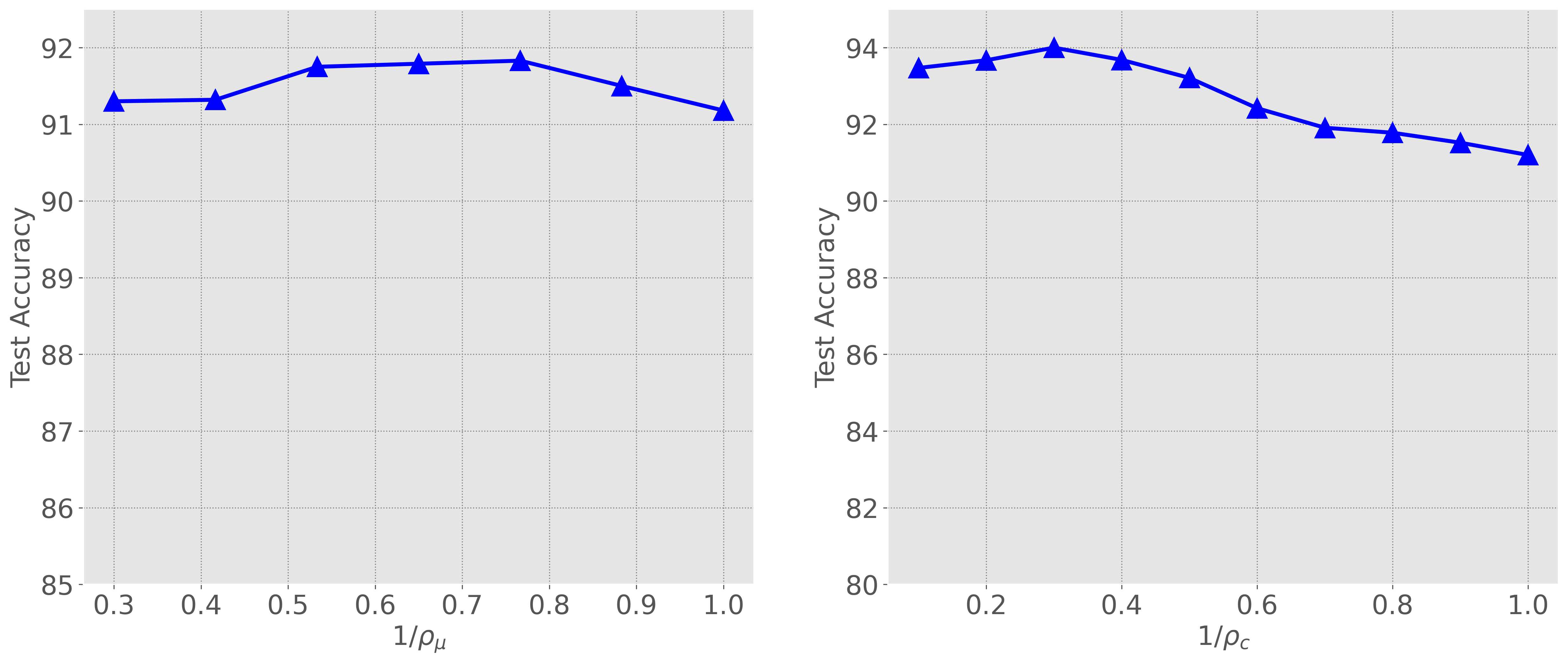}
    \caption{Dynamic DP-SGD performance is robust to  $\rho_{\mu}$ and $\rho_c$.}
    \label{fig:decay_rate_search}
\end{figure}

We then test the robustness of dynamic DP-SGD performance to different values of $\rho_{\mu}$ and $\rho_c$ as shown in Fig.~\ref{fig:decay_rate_search}.
We use grid search to demonstrate the impact of these parameters on model performance. The dynamic method can consistently improve model performance across a wide range.

 \subsection{Exact Privacy Cost}
\begin{figure}[h]
    \centering
    \includegraphics[width = 0.25\textwidth]{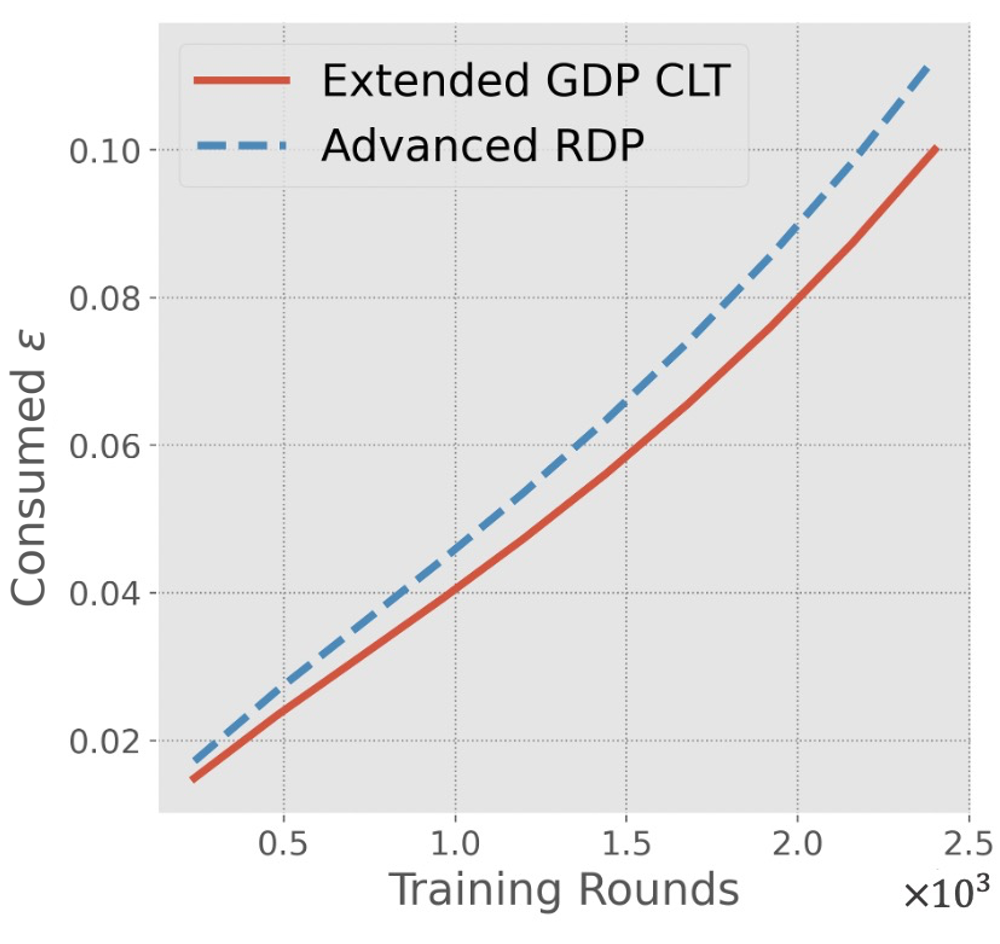}
    \caption{Upper bound (advanced RDP) and lower bound (extended GDP CLT) of the dynamic DP-SGD privacy  cost curves. The true privacy cost  must lie somewhere between these two limits. The result of extended CLT for GDP is reasonable because the privacy cost differences between these two limits are small, especially  in the high privacy protection region.}
    \label{fig:rdp_gdp_comp}
\end{figure}

Though~\cite{dong2021gaussian} has shown  that the  GDP CLT approximate the true privacy cost with negligible error,
\citet{gopi2021numerical} recently discovers that the GDP CLT may underestimate the privacy cost.
The RDP accountant~\citep{wang2019subsampled}, on the other hand, overestimates the true cost.
To evaluate the exact privacy cost, we plot the privacy cost curves for both the proposed extended GDP CLT and RDP in Fig.~\ref{fig:rdp_gdp_comp}. Specifically, we set a target training round and conduct privacy accounting for the dynamic DP-SGD, based on the extended  GDP CLT as well as RDP\footnote{For RDP, we use autodp library by \cite{autodp} for DP accounting.}~\citep{balle2020privacy}.
Consequently,  the true privacy cost curve must lie somewhere between these two limits. The result of extended GDP CLT is reasonable because the privacy cost differences between these two limits are small, particularly in the high privacy protection region.

It is worth noting that exact accounting can be performed using the recently proposed methods by~\cite{gopi2021numerical} and~\cite{zhu2021optimal}, separately. However, because they both require numerical computation, given a total privacy budget, the computation of different noise power for  the mechanism of each step  becomes much involved.

% \textbf{IMDB and NAME} 
% Experiments on IMDB and NAME is shown in Table \ref{imdb_res} and \ref{name_res}. We observed a larger performance degradation under NLP tasks. In this case, we apply larger $\epsilon$.

% To train language models, an important component is embedding layer, which is a large part from parameters' point of view and is quite sensitive to noise. Thus making training language models more difficult. We compare the result of training with frozen pretrained embedding layer and with tunable pretrained embedding layer.

% \subsection{Convergence Analysis}
% In this section, we discuss the impact of convergence of our proposed dynamic noise. The sensitivity of gradients during training is the clipping value we set. However, when training with static noise across iterations, we observe that the gradient norm tends to diverge, this is due to the additive noise upon the gradient. By allocating noise, we alleviate this  divergence and turn it to converge.
\begin{table}[htb!]
\centering
\caption{CNN on  MNIST and FashionMNIST datasets.}
\begin{adjustbox}{width=0.48\textwidth}
\hfill{}\begin{tabular}{|c|c|c|c|c|c|c|c|}
\hline 
% \multirow{2}{*}{} 
\multirow{2}{*}{DP Accountant}&\multirow{2}{*}{Dynamic Noise} & \multicolumn{3}{c|}{ MNIST} & \multicolumn{3}{c|}{{FashionMNIST}} 
\tabularnewline
\cline{3-5} \cline{6-8}
 && {$\epsilon$ = 0.4} & {$\epsilon$ =0.6}& {$\epsilon$ =1.2} & {$\epsilon$ =0.4} & {$\epsilon$ =1.2}& {$\epsilon$ =2.0}\tabularnewline
\hline 
\hline 
{Non-private} &{-}& \multicolumn{3}{c|}{{98.83}}  & \multicolumn{3}{c|}{{87.92}}   \tabularnewline
\hline 
\hline 
{$\rho$-zCDP-SGD} &Noise Decay& {10.28} & {10.12} & {65.33} & {9.86} & 63.30 & 72.18
\tabularnewline
\hline 
{tCDP-SGD } &Noise Decay& {26.93} & {83.28} & {92.60} & {53.69} & 76.48 & 77.58
\tabularnewline
%{RDP} &Dynamic DP& {86.43} & {91.80} & {94.42} & {-} & 78.70 & 80.74 \tabularnewline
\hline
{Vanilla  DP-SGD} &-& {91.18} & {93.80} & {95.50} & {76.77} & 80.45 & 82.55 \tabularnewline
\hline 
\hline 
\multirow{3}{*}{Ours}
&Growing-$\mu_t$  & {91.67} & {94.49} & 96.06& {77.81} & 80.95  & 83.10  \tabularnewline
\cline{2-8}
%\cline{3-8}
 & Sensitivity Decay & {93.95} & 95.17 & 96.17 & {78.11} & 82.83  & 83.64  \tabularnewline

% \hline 
\cline{2-8}
&Dynamic DP & {\textbf{94.35}} & {\textbf{95.21}} & {\textbf{96.34}} & {\textbf{78.50}} & {\textbf{83.22}} & \textbf{83.81} \tabularnewline
\hline 

\end{tabular}\hfill{}
\end{adjustbox}
\label{mnist_res}
\end{table}

\begin{table}[htb!]
\centering
\caption{MLP on the IMDB dataset.}
\begin{adjustbox}{width=0.48\textwidth}
\hfill{}\begin{tabular}{|c|c|c|c|c|c|c|}
\hline 
% \multirow{2}{*}{} 
\multirow{2}{*}{DP Accountant}&\multirow{2}{*}{Dynamic Noise} & \multicolumn{5}{c|}{IMDB} 
\tabularnewline
\cline{3-7} 
 && {$\epsilon$ = 0.5} & {$\epsilon$ =1}& {$\epsilon$ =3} & {$\epsilon$ =6} & {$\epsilon$ =9}\tabularnewline
\hline 
\hline 
{Non-private} &{-}& \multicolumn{5}{c|}{{82.85}}  \tabularnewline
\hline 
\hline
% {$\rho$-zCDP(RF)} &Noise Decay& {-} & {-} & {-} & {-} & -  \tabularnewline
% \hline 

{tCDP-SGD } &Noise Decay&56.67 &{58.24} & {62.15} & {65.88} & {70.16} 
\tabularnewline
\hline 
{Vanilla DP-SGD} &-& {63.62} & 69.71 & {75.64} & {77.75} & 78.60  \tabularnewline
\hline 
\hline 
\multirow{3}{*}{Ours}
&Growing-$\mu_t$ & {64.92} & {69.85} & 76.00& {78.16} & 78.56  \tabularnewline
\cline{2-7}
%\cline{3-8}
 & Sensitivity Decay & {65.44} & {70.25} & 76.23 & {78.47} & 79.42    \tabularnewline

% \hline 
\cline{2-7}
&Dynamic GDP & {\textbf{65.63}} & {\textbf{70.77}} & {\textbf{76.64}} & {\textbf{78.61}} & {\textbf{79.61}}  \tabularnewline
\hline 

\end{tabular}\hfill{}
\end{adjustbox}
\label{imdb_res}
\end{table}

\begin{table}[htb!]
\centering
\caption{LSTM on NAME dataset.}
\begin{adjustbox}{width=0.48\textwidth}
\hfill{}\begin{tabular}{|c|c|c|c|c|c|}
\hline 
% \multirow{2}{*}{} 
\multirow{2}{*}{DP Framework}&\multirow{2}{*}{Dynamic Noise} & \multicolumn{4}{c|}{ NAME} 
\tabularnewline
\cline{3-6} 
 && {$\epsilon$ = 1} & {$\epsilon$ =2}& {$\epsilon$ =4} & {$\epsilon$ =8} \tabularnewline
\hline 
\hline 
{Non-private} &{-}& \multicolumn{4}{c|}{{80.14}}  \tabularnewline
\hline 
\hline
% {$\rho$-zCDP(RF)} &Noise Decay& {-} & {-} & {-} & {-}  \tabularnewline
% \hline 

{tCDP-SGD } &Noise Decay&48.13 &{51.20} & {57.67} & {68.53}
\tabularnewline
\hline 

{Vanilla DP-SGD} &-& {62.71} & {69.64} & {73.04} & {74.15}   \tabularnewline
\hline 
\hline 
\multirow{3}{*}{Extended CLT for GDP (Ours)}
&growing-$\mu_t$ & {64.10} & {71.25} & 74.68& {75.50}   \tabularnewline
\cline{2-6}
%\cline{3-8}
 & Sensitivity Decay & {66.66} & {71.78} & {73.67} & {74.78}     \tabularnewline

% \hline 
\cline{2-6}
&Dynamic DP & {\textbf{67.30}} & {\textbf{72.01}} & {\textbf{75.03}} & {\textbf{75.75}}   \tabularnewline
\hline 

\end{tabular}\hfill{}
\end{adjustbox}
\label{name_res}
\end{table}

\begin{table}[htb!]
\centering
\caption{Federated learning on InfiniteMNIST dataset.}
\begin{adjustbox}{width=0.48\textwidth}
\hfill{}\begin{tabular}{|c|c|c|c|c|c|c|c|}
\hline 
% \multirow{2}{*}{} 
\multirow{2}{*}{DP Framework}&\multirow{2}{*}{Dynamic Noise} & \multicolumn{3}{c|}{MNIST-250K} & \multicolumn{3}{c|}{{MNIST-500K}} 
\tabularnewline
\cline{3-5} \cline{6-8}
 && {$\epsilon$ = 0.1} & {$\epsilon$ =0.4}& {$\epsilon$ =1} & {$\epsilon$ =0.1} & {$\epsilon$ =0.4}& {$\epsilon$ = 1}\tabularnewline
\hline 
\hline 
{Non-private} &{-}& \multicolumn{3}{c|}{{98.89}}  & \multicolumn{3}{c|}{{98.96}}   \tabularnewline
\hline 
{Vanilla DP-SGD} &-& {93.47} & {95.71} & {96.02} & {94.82} & {96.55} & {96.89} \tabularnewline
\hline 
\hline 
\multirow{3}{*}{Ours}
&growing-$\mu_t$ & {93.75} & {95.93} & 96.13& {95.65} & {96.76} & 97.05  \tabularnewline
\cline{2-8}
%\cline{3-8}
 & Sensitivity Decay & {94.46} & 95.90 & 96.40 & {95.75} & 96.83 & 97.06 \tabularnewline

% \hline 
\cline{2-8}
&Dynamic DP & \textbf{94.72} & {\textbf{96.00}} & {\textbf{96.55}} & \textbf{95.88} & {\textbf{96.95}} & {\textbf{97.22}} \tabularnewline
\hline 

\end{tabular}\hfill{}
\end{adjustbox}
\label{FL_res}
\end{table}

\begin{table}[htb!]
\centering
\caption{VGG-11 on chest radiographs.}
\begin{adjustbox}{width=0.4\textwidth}
\hfill{}\begin{tabular}{|c|c|c|c|c|}
\hline 
% \multirow{2}{*}{} 
\multirow{2}{*}{DP Framework}&\multirow{2}{*}{Dynamic Noise} & \multicolumn{3}{c|}{ Chest Xray} 
\tabularnewline
\cline{3-5} 
 && {$\epsilon$ = 0.5} & {$\epsilon$ =1}& {$\epsilon$ =1.5}  \tabularnewline
\hline 
\hline 
{Non-private} &{-}& \multicolumn{3}{c|}{{97.02}}  \tabularnewline
\hline 
\hline
% {$\rho$-zCDP(RF)} &Noise Decay& {-} & {-} & {-} & {-}  \tabularnewline
% \hline 

{Vanilla DP-SGD} &-& {92.24} & {93.18} & {93.44}   \tabularnewline
\hline 
\hline 
\multirow{3}{*}{Ours}
&growing-$\mu_t$ & {92.60} & {93.65} & 94.01   \tabularnewline
\cline{2-5}
%\cline{3-8}
 & Sensitivity Decay & {92.89} & {93.78} & {94.24}     \tabularnewline

% \hline 
\cline{2-5}
&Dynamic DP & {\textbf{93.30}} & {\textbf{94.01}} & {\textbf{94.43}}  \tabularnewline
\hline 

\end{tabular}\hfill{}
\end{adjustbox}
\label{chest_res}
\end{table}

% We compare our proposed dynamic DP with poisson subsampled RDP \citep{zhu2019poission} as RDP also supports analysis of dynamic noise power. 
% The comparison of privacy budget consumption curve is shown in Figure \ref{fig:rdp_gdp_comp}. With fixed $\delta=10^{-5}$, we set a target $\epsilon=0.8$ and calculate noise power sequence by GDP algorithm in a static or dynamic manner, then we  apply GDP and RDP framework to analyze the noise power sequence to generate privacy budget consumption curve. We first compare static GDP and RDP, as we have illustrated, GDP achieves tighter bound of privacy loss estimation thus a smaller $\epsilon$ is obtained. Similarly, with dynamic noise power, Dynamic-GDP maintains a smaller $\epsilon$. In the begining, with large noise power, Dynamic GDP performs slightly better than RDP. Nevertheless, the gap between Dynamic GDP and RDP tends to broad as training goes on.
% Moreover, in terms of computation overhead, thousands of training iterations with dynamic noise introduce the same number of types of mechanism in the RDP composition. This requires much more computation budget than dynamic GDP as RDP involves divergence computing and requires to find a global $\alpha$ for the composed mechanism. When there are thousands of mechanism with different noise power, $\alpha$ computing becomes quite slow. Nevertheless, this is not a problem for dynamic GDP, our composition formula is easy to implement and the computation time is negligible comparing with RDP. 

\section{Theoretical Analysis}
\label{sec:analysis}
In this section, we theoretically examine the impact of varying DP costs  on DP-SGD utility guarantees. To begin with, we state a general form of dynamic DP-SGD where $C_t$ and $\sigma_t$ can be arbitrary predefined sequences in Algorithm~\ref{alg:general-DPSGD}.
\begin{algorithm}[h]
%\begin{algorithm}[htbp]
\caption{General Dynamic DP-SGD Algorithm}
\label{alg:general-DPSGD}
\begin{algorithmic}[1]
\REQUIRE  DP budget $(\epsilon,\delta)$, sampling rate $p$, clipping threshold $C_1, ..., C_T$, and noise power $\widetilde \sigma\cdot \widetilde \sigma_1, \widetilde\sigma\cdot \widetilde \sigma_2, ...,  \widetilde \sigma\cdot\widetilde\sigma_T $.
\vspace{-12pt}
% \STATE Compute $\mu_{\text{tot}}$ accordin to (\ref{compute_mu}) \Note{SC: do we need this step?}
\FOR{$t=1, \ldots, T$}
\STATE 
Sample $ X_t\in X$ with sampling rate $p$ and sample noise $\xi_{t}\sim \mathcal N(0, \widetilde\sigma^2 I)$. 
\STATE Compute:
    $\theta_{t+1} =  \theta_{t} - 
    \frac{\eta }{p\left|X\right|}\big[\sigma_t \xi_{t}+$
    $ 
    \sum_{x\in X_{ t}}\text{CL}\left(g_x;C_t\right)
    \big]$ with $\text{CL}(\cdot)$ in (\ref{clipping})
\ENDFOR
\end{algorithmic}
\end{algorithm}
% \vspace{-2cm}
Compared with Algorithm 2, aside from allowing arbitrary sequences $C_0, C_1, ..., C_T$ and $\sigma_1, \sigma_2, ..., \sigma_T$, we denote $\sigma_t = \widetilde\sigma \cdot \widetilde\sigma_t $ for the convenience of analysis.  Besides, in step 3 of Algorithm~\ref{alg:general-DPSGD}, we use $p|X|$ to replace $|X_t|$ with the intuition being $\mathbb E[|X_t|] = p|X|$, where $p$ is the sampling rate. Such a replacement can also be found in \citet{DBLP:conf/ccs/AbadiCGMMT016}, which is one of the most well-known version of DP-SGD. The main advantage of this form in our situation  is that it simplifies  the utility analysis by removing the need to calculate the first and second moments of $\frac{1}{|X_t|}$, which take complicated forms. To proceed with utility analysis,  we first state a 
corollary of Theorem~\ref{MU-CLT} regarding the requirement of $\widetilde \sigma$ with proof provided in Appendix~\ref{app:proof-corrollary1}.
\begin{corollary}\label{thm: dp_alg3}
 Algorithm 3 satisfies $G_{\mu_{\text{tot}}}$-DP if
 {\small
 \begin{align}\label{eq: sigma_tot}
    \widetilde \sigma = p \frac{1}{\mu_{\text{tot}}}\sqrt{2\sum_{t=1}^T \frac{C_t^2}{\widetilde \sigma_t^2}},
\end{align}
}%
and $\frac{C_t}{\widetilde \sigma\widetilde \sigma_{t}}\leq 1, \forall t \in [T]$.
\end{corollary}
Corollary \ref{thm: dp_alg3} is a direct implication of Theorem \ref{MU-CLT} with a mild constraint $\frac{C_t}{\widetilde \sigma_{t}\widetilde \sigma}=\mu_t \leq 1$. Note that $\mu_t \leq 1$ is guaranteed for strong privacy as  $\mu_t\ll\mu_{\text{tot}}$, and $\mu_{\text{tot}} = 1$ is just
borderline private pointed out by \citet{dong2021gaussian}.
The smaller $\mu_{\text{tot}}$, the more privacy it preserves.

Denote the objective function  as $ L(\theta) = \frac{1}{N}\sum_{i=1}^N f(\theta, x_i)$ and the clipped stochastic gradient at step $t$ as $g_t = \frac{1}{p\left|X\right|}\big[
    \sum_{x\in X_{ t}}\text{CL}\left(g_x;C_t\right)
    \big]$ where $g_x \triangleq \frac{\partial f(\theta_t, x) }{\partial \theta_t} $.  
    Assume function $L(\theta)$ is $Q$-smooth, i.e., $L(\theta') - L(\theta) \leq  \langle \nabla L(\theta), \theta'-\theta\rangle + \frac{Q}{2}\|\theta-\theta'\|^2$), and each per-sample gradient is upper-bounded, i.e. $\|g_x\| \leq G$.  We prove the following theorem to show the privacy-utility trade-off of Algorithm~\ref{alg:general-DPSGD} with details in Appendix~\ref{appendix:thm-utl_alg3}.
\begin{theorem}\label{thm: utl_alg3}
The dynamic DP-SGD (Algorithm 3) 
satisfies the following utility guarantee
{\small
\begin{align}\label{eq: sgd_bound}
      & \mathbb E\left[ \frac{1}{T} \sum_{t=1}^T \| \nabla   L(\theta_{t})\|^2 \right] \nonumber \\
      & \leq   \frac{1}{\eta T}  \mathbb  E[L(\theta_{1})] - L(\theta_{T+1})] 
     +   \frac{Q}{2} \frac{\eta}{T}  \sum_{t=1}^T  \mathbb E[ \| g_{t}\|^2]  \nonumber \\ 
      & + \underbrace{\mathbb E \left[\frac{1}{T} \sum_{t=1}^T \| \nabla   L(\theta_{t})\| P_t(C_t) G \right]}_{\triangleq\text{ bias term}}+ \frac{Q}{2}  \frac{\eta}{p^2N^2}\underbrace{ \frac{1}{T }\sum_{t=1}^T \widetilde\sigma^2\widetilde\sigma_t^2  }_{ \triangleq D_1},
\end{align}
}%
where $P_t(C_t)$ is the probability of a per-sample gradient being clipped with threshold $C_t$ at $\theta_t$, and the expectations are taken over all randomness including gradient sampling and noise sampling. Assume $\sigma_t =  \Theta(1)$ and $C_t = \Theta(1)$, after setting $\eta T = N \mu_{\text{tot}}$, $\eta = \frac{1}{N\mu_{\text{tot}}}$, and substituting $\widetilde \sigma$ from Corollary \ref{thm: dp_alg3} into Eq. (\ref{eq: sgd_bound}), we have $$\mathbb E\left[ \frac{1}{T} \sum_{t=1}^T \| \nabla   L(\theta_{t})\|^2 \right]  =  O\left(\frac{1}{N\mu_{\text{tot}}}\right) + \text{bias term}.$$ %{\red[not sure what the lat if means?][the bias is the same as the bias in (16)?]}
\end{theorem}

\textbf{Remark on convergence rate:} The derived convergence rate is $O({1}/{(N\mu_{\text{tot}}}))$ ignoring the non-vanishing bias term. This rate is in the same polynomial order as existing analyses, e.g., $\tilde O(1/(N\epsilon))$ rate in Theorem 5 of \citet{zhang2017efficient}. The non-vanishing bias term caused by gradient clipping is unavoidable since it is proven in \citet{song2020characterizing} that DP-SGD with clipping suffers a constant regret in the worst case. A similar bound  is also observed in \citet{DBLP:conf/nips/ChenWH20}, in which the impact of gradient clipping on symmetric gradient distibutions is studied. The difference is that our bound is based on Gaussian DP and the bias term is expressed in a simpler form applicable for arbitrary distributions. As will be evident later, this simpler form provides intuitions on why clipping threshold and noise should be adjusted dynamically.

\textbf{Remark on gradient clipping $C_t$:} The clipping operation will cause estimation bias of the true gradient, which could lead to a constant regret for the algorithm. This fact is reflected by the $bias$ term on the r.h.s. of \eqref{eq: sgd_bound}. And it can be seen from $P_t(C_t)$ (defined as the probability of a per-sample gradient being clipped with threshold $C_t$ at $\theta_t$) that in general, the $bias$ term will be small if $C_t$'s are large and vise versa, given the same distribution of gradients. This means that if we use large clipping thresholds, we will reduce the $bias$ term, but at the expense of a larger $D_1$ as indicated by Corollary \ref{thm: dp_alg3} and \eqref{eq: sgd_bound}. Thus, one should choose a sequence of $C_t$ that could better balance the bias term and the variance term $D_1$. However, an optimal solution requires knowing the distribution of the stochastic gradient which is nearly impossible in practice. Choosing $C_t$ is more of an art than a science, according to the existing literature. One notable practical good choice is to keep $P_t(C_t)$ roughly constant, as proposed in \cite{andrew2019differentially}. Under situations when the gradients are decreasing across iterations, this choice implies $C_t$ should be decreasing to keep $P_t(C_t)$ a constant (see Figure 4 in \cite{andrew2019differentially}). This supports the sensitivity decay method in our Algorithm 2.
%We first notice that the LHS of \eqref{eq: sgd_bound} is not a traditional convergence measure like average of squared norm of gradients, this is  because the estimation bias in gradient estimator caused by gradient clipping. There are existing ways to relate this to traditional convergence measures by separating the bias using clipping probability \citep{zhang2019gradient} or Wasserstein distance between distributions \citep{DBLP:conf/nips/ChenWH20}. We will study this in Appendix 

\textbf{Remark on term $D_1$:} The first two terms on the RHS of \eqref{eq: sgd_bound} are standard in SGD analysis. We now focus on the term $D_1$ caused by privacy noise, which becomes 
\begin{align}\label{eq:dp_variance-D1} 
    p^2 \frac{2}{\mu_{\text{tot}}^2} \frac{1}{T }\sum_{t=1}^T \widetilde \sigma_t^2 {\sum_{t=1}^T \frac{C_t^2}{\widetilde \sigma_t^2}}
\end{align}
after substituting $\widetilde \sigma$ from Corollary~\ref{thm: dp_alg3}. One can readily notice that given the sequence of $C_t$, this term can be minimized by choosing $\widetilde \sigma_t$. This problem also appeared in \citet{AdaClip} and \citet{wu2021adaptive}. It turns out that minimizing \eqref{eq:dp_variance-D1} w.r.t the sequence $\sigma_t$ admits an optimal solution $\widetilde \sigma_t  = \sqrt{C_t}$ (see Appendix \ref{app: opt_sol} for proof). Combining with the fact that $\mu_t = \frac{C_t}{\widetilde \sigma_t\widetilde \sigma}$  indicates $\mu_t \propto \sqrt{C_t}$, which gives a theoretical way to allocate noise given a sequence of clipping thresholds and support that we should change $\mu_t$ in different iterations. Yet, this particular choice of $\sigma_t$ is based on the very simplified assumption that may not reflect the whole picture of training neural nets. For example, it is widely known that the loss of training neural nets usually have different curvatures across different regions reachable by different iterations. If we take into this consideration and replace the Q-smooth assumption by $\mathbb E[L(\theta_{t+1})] - L(\theta) \leq  \langle \nabla L(\theta_t), \mathbb E[\theta_{t+1}-\theta_t] \rangle + \frac{Q_t}{2} \mathbb E[\|\theta_{t+1}-\theta_t\|^2]$ in Theorem~\ref{thm: utl_alg3}, we will reach a conclusion that $\sigma_t  = \sqrt{C_t}Q_t^{1/4}$, which indicates the allocation of privacy noise should also consider the local curvature in different iterations unknown before training (see Appendix \ref{app: local_cur} for a  detailed statement and proof). We thus believe that a good choice of $\sigma_t$ and $C_t$ should be based more on empirical performance of a particular choice, like the one provided in Section~\ref{sec:experiments}. Meanwhile, we hope our theoretical analysis can provide some insights about understanding this problem and inspire future explorations.

\section{Related Work}
% The following section summarizes related work on algorithm design and DP accounting for DP-SGD.

\noindent\textbf{{DP-SGD Algorithm}}
To improve the model's accuracy, previous work has concentrated on designing variations of DP-SGD  by estimating the clipping bound and minimizing the bias introduced by gradient clipping. More precisely, \citet{DBLP:conf/ccs/AbadiCGMMT016} propose norm clipping and per-layer clipping, both of which select clipping values based on gradient differences between different layers.
\citet{AdaClip} study AdaClip, a coordinate-wise clipping method that could reduce the total amount of noise required.
\citet{Quantile} introduce  gradient clipping  based on  the quantile statistics of the gradient, which requires additional DP cost to protect those quantiles.
Recently, \citet{DBLP:conf/nips/ChenWH20} analyze the bias introduced by the gradient clipping operation and propose a method for reducing the bias error by first adding noise before clipping.
In the meantime of this paper, \citet{wu2021adaptive} propose adaptive DP version of SGD where the random noise added to the gradient is optimally adapted to the stepsize. This method, however, is inapplicable for those with stepsizes that being updated on the fly, such as AdaGrad and Adam, as stated in the paper.
It's worth noting that the proposed dynamic DP-SGD in this paper is compatible with the methods in~\citet{Quantile,DBLP:conf/nips/ChenWH20,wu2021adaptive} and can be used in tandem to investigate accuracy improvement but involve more details and is one potential avenue of future work.

Furthermore,
\citet{yu2019differentially} provide a means of reducing noise variance during the DP-SGD process, thereby improving model performance; and  \citet{zhang2021adaptive} analyze the DP cost for the same method using the z-CDP privacy accounting.
Due to the loose DP accountings, these methods have a large performance gap in the high privacy guaranteed region. As demonstrated  in the experiments, the proposed dynamic DP-SGD improves these results significantly due to the dynamic clipping operation and tight DP composition. Recently, \citet{zhou2020private, hetighter} study the relationships between generalization
and privacy  private learning algorithm.

\section{Conclusions}
In this paper, 
we have developed the  dynamic differentially-private stochastic gradient descent (dynamic DP-SGD) optimizer, which has varying clipping values and noise powers across the update. The dynamic privacy cost is tightly accounted  by the extended central limit theorem of Gaussian differential privacy, allowing dynamic noise to be calibrated for each individual training step within a predefined privacy budget.
In contrast to the vanilla DP-SGD, we are able to reduce the noise term in the utility upper bound without compromising privacy 
as the insight gained according to our theoretical analysis.
Extensive testing on a variety of datasets and models demonstrates that the dynamic DP-SGD consistently and clearly outperforms existing methods especially in the strong privacy region.

% For the future work, we intend to investigate how to leverage the  recently achievement of numerical composition methods \citep{gopi2021numerical, zhu2021optimal} to calibrate the noise power and clipping value for each individual training step of dynamic DP-SGD.
% We ran experiments on a variety of datasets and models, and the results show that when combined with the proposed accounting approach, dynamic DP-SGD clearly outperforms the competitions.

% Furthermore, our method is consistent with other recent advances in DP-SGD, allowing us to improve the model's accuracy even further.

\subsubsection*{Acknowledgments}
We would like to thank Yuxiang Wang for his helpful discussion.

\bibliography{iclr2022_conference}

\begin{thebibliography}{28}
\providecommand{\natexlab}[1]{#1}
\providecommand{\url}[1]{\texttt{#1}}
\expandafter\ifx\csname urlstyle\endcsname\relax
  \providecommand{\doi}[1]{doi: #1}\else
  \providecommand{\doi}{doi: \begingroup \urlstyle{rm}\Url}\fi

\bibitem[Abadi et~al.(2016)Abadi, Chu, Goodfellow, McMahan, Mironov, Talwar,
  and Zhang]{DBLP:conf/ccs/AbadiCGMMT016}
Mart{\'{\i}}n Abadi, Andy Chu, Ian~J. Goodfellow, H.~Brendan McMahan, Ilya
  Mironov, Kunal Talwar, and Li~Zhang.
\newblock Deep learning with differential privacy.
\newblock In Edgar~R. Weippl, Stefan Katzenbeisser, Christopher Kruegel,
  Andrew~C. Myers, and Shai Halevi (eds.), \emph{Proceedings of the 2016 {ACM}
  {SIGSAC} Conference on Computer and Communications Security, Vienna, Austria,
  October 24-28, 2016}, pp.\  308--318. {ACM}, 2016.

\bibitem[Andrew et~al.(2021)Andrew, Thakkar, McMahan, and
  Ramaswamy]{andrew2019differentially}
Galen Andrew, Om~Thakkar, H~Brendan McMahan, and Swaroop Ramaswamy.
\newblock Differentially private learning with adaptive clipping.
\newblock \emph{Advances in Neural Information Processing Systems 34
  pre-proceedings}, 2021.

\bibitem[Balle et~al.(2020)Balle, Barthe, and Gaboardi]{balle2020privacy}
Borja Balle, Gilles Barthe, and Marco Gaboardi.
\newblock Privacy profiles and amplification by subsampling.
\newblock \emph{Journal of Privacy and Confidentiality}, 10\penalty0 (1), 2020.

\bibitem[Bottou et~al.(2018)Bottou, Curtis, and
  Nocedal]{bottou2018optimization}
L{\'e}on Bottou, Frank~E Curtis, and Jorge Nocedal.
\newblock Optimization methods for large-scale machine learning.
\newblock \emph{Siam Review}, 60\penalty0 (2):\penalty0 223--311, 2018.

\bibitem[Bu et~al.(2020)Bu, Dong, Long, and Su]{bu2019deep}
Zhiqi Bu, Jinshuo Dong, Qi~Long, and Weijie~J. Su.
\newblock Deep learning with {G}aussian differential privacy.
\newblock \emph{Harvard data science review}, 2020 23, 2020.

\bibitem[Chen et~al.(2020)Chen, Wu, and Hong]{DBLP:conf/nips/ChenWH20}
Xiangyi Chen, Zhiwei~Steven Wu, and Mingyi Hong.
\newblock Understanding gradient clipping in private {SGD:} {A} geometric
  perspective.
\newblock In Hugo Larochelle, Marc'Aurelio Ranzato, Raia Hadsell,
  Maria{-}Florina Balcan, and Hsuan{-}Tien Lin (eds.), \emph{Advances in Neural
  Information Processing Systems 33: Annual Conference on Neural Information
  Processing Systems 2020, NeurIPS 2020, December 6-12, 2020, virtual}, 2020.

\bibitem[Dong et~al.(2019)Dong, Roth, and Su]{dong2021gaussian}
Jinshuo Dong, Aaron Roth, and Weijie~J Su.
\newblock Gaussian differential privacy.
\newblock \emph{arXiv preprint arXiv:1905.02383}, 2019.

\bibitem[Dwork(2006)]{DBLP:conf/icalp/Dwork06}
Cynthia Dwork.
\newblock Differential privacy.
\newblock In Michele Bugliesi, Bart Preneel, Vladimiro Sassone, and Ingo
  Wegener (eds.), \emph{Automata, Languages and Programming, 33rd International
  Colloquium, {ICALP} 2006, Venice, Italy, July 10-14, 2006, Proceedings, Part
  {II}}, volume 4052 of \emph{Lecture Notes in Computer Science}, pp.\  1--12.
  Springer, 2006.

\bibitem[Dwork et~al.(2014)Dwork, Roth, et~al.]{dwork2014algorithmic}
Cynthia Dwork, Aaron Roth, et~al.
\newblock The algorithmic foundations of differential privacy.
\newblock \emph{Found. Trends Theor. Comput. Sci.}, 9\penalty0 (3-4):\penalty0
  211--407, 2014.

\bibitem[Gopi et~al.(2021)Gopi, Lee, and Wutschitz]{gopi2021numerical}
Sivakanth Gopi, Yin~Tat Lee, and Lukas Wutschitz.
\newblock Numerical composition of differential privacy.
\newblock \emph{arXiv preprint arXiv:2106.02848}, 2021.

\bibitem[He et~al.()He, Wang, and Tao]{hetighter}
Fengxiang He, Bohan Wang, and Dacheng Tao.
\newblock Tighter generalization bounds for iterative privacy-preserving
  algorithms.
\newblock In \emph{the 37th Conference on Uncertainty in Artificial
  Intelligence (UAI 2021)}.

\bibitem[Kermany et~al.(2018)Kermany, Goldbaum, Cai, Valentim, Liang, Baxter,
  McKeown, Yang, Wu, Yan, et~al.]{kermany2018identifying}
Daniel~S Kermany, Michael Goldbaum, Wenjia Cai, Carolina~CS Valentim, Huiying
  Liang, Sally~L Baxter, Alex McKeown, Ge~Yang, Xiaokang Wu, Fangbing Yan,
  et~al.
\newblock Identifying medical diagnoses and treatable diseases by image-based
  deep learning.
\newblock \emph{Cell}, 172\penalty0 (5):\penalty0 1122--1131, 2018.

\bibitem[Lecun \& Bottou(1998)Lecun and Bottou]{mnist}
Y.~Lecun and L.~Bottou.
\newblock Gradient-based learning applied to document recognition.
\newblock \emph{Proceedings of the IEEE}, 86\penalty0 (11):\penalty0
  2278--2324, 1998.

\bibitem[Pichapati et~al.(2019)Pichapati, Suresh, Yu, Reddi, and
  Kumar]{AdaClip}
Venkatadheeraj Pichapati, Ananda~Theertha Suresh, Felix~X. Yu, Sashank~J.
  Reddi, and Sanjiv Kumar.
\newblock Adaclip: Adaptive clipping for private sgd.
\newblock \emph{ArXiv}, abs/1908.07643, 2019.

\bibitem[Song et~al.(2020)Song, Thakkar, and Thakurta]{song2020characterizing}
Shuang Song, Om~Thakkar, and Abhradeep Thakurta.
\newblock Characterizing private clipped gradient descent on convex generalized
  linear problems.
\newblock \emph{arXiv preprint arXiv:2006.06783}, 2020.

\bibitem[Thakkar et~al.(2019)Thakkar, Andrew, and McMahan]{Quantile}
Om~Thakkar, Galen Andrew, and H.~Brendan McMahan.
\newblock Differentially private learning with adaptive clipping.
\newblock \emph{CoRR}, abs/1905.03871, 2019.

\bibitem[Wang et~al.(2019)Wang, Balle, and Kasiviswanathan]{wang2019subsampled}
Yu-Xiang Wang, Borja Balle, and Shiva~Prasad Kasiviswanathan.
\newblock Subsampled {R}{\'e}nyi differential privacy and analytical moments
  accountant.
\newblock In \emph{The 22nd International Conference on Artificial Intelligence
  and Statistics}, pp.\  1226--1235. PMLR, 2019.

\bibitem[Wang()]{autodp}
Yuxiang Wang.
\newblock autodp: A flexible and easy-to-use package for differential privacy.
\newblock \url{https://github.com/yuxiangw/autodp}.
\newblock Accessed: 2021-09-30.

\bibitem[Wu et~al.(2021)Wu, Wang, Cristali, Gu, and Willett]{wu2021adaptive}
Xiaoxia Wu, Lingxiao Wang, Irina Cristali, Quanquan Gu, and Rebecca Willett.
\newblock Adaptive differentially private empirical risk minimization.
\newblock \emph{arXiv preprint arXiv:2110.07435}, 2021.

\bibitem[Yu et~al.(2019)Yu, Liu, Pu, Gursoy, and Truex]{yu2019differentially}
Lei Yu, Ling Liu, Calton Pu, Mehmet~Emre Gursoy, and Stacey Truex.
\newblock Differentially private model publishing for deep learning.
\newblock In \emph{2019 IEEE Symposium on Security and Privacy (SP)}, pp.\
  332--349. IEEE, 2019.

\bibitem[Zhang et~al.(2017)Zhang, Zheng, Mou, and Wang]{zhang2017efficient}
Jiaqi Zhang, Kai Zheng, Wenlong Mou, and Liwei Wang.
\newblock Efficient private erm for smooth objectives.
\newblock \emph{arXiv preprint arXiv:1703.09947}, 2017.

\bibitem[Zhang et~al.(2021{\natexlab{a}})Zhang, Ding, Wu, Wong, Nguyen, and
  Pan]{CDP}
Xinyue Zhang, Jiahao Ding, Maoqiang Wu, Stephen T.~C. Wong, Hien~Van Nguyen,
  and Miao Pan.
\newblock Adaptive privacy preserving deep learning algorithms for medical
  data.
\newblock In \emph{{IEEE} Winter Conference on Applications of Computer Vision,
  {WACV} 2021, Waikoloa, HI, USA, January 3-8, 2021}, pp.\  1168--1177. {IEEE},
  2021{\natexlab{a}}.

\bibitem[Zhang et~al.(2021{\natexlab{b}})Zhang, Ding, Wu, Wong, Van~Nguyen, and
  Pan]{zhang2021adaptive}
Xinyue Zhang, Jiahao Ding, Maoqiang Wu, Stephen~TC Wong, Hien Van~Nguyen, and
  Miao Pan.
\newblock Adaptive privacy preserving deep learning algorithms for medical
  data.
\newblock In \emph{Proceedings of the IEEE/CVF Winter Conference on
  Applications of Computer Vision}, pp.\  1169--1178, 2021{\natexlab{b}}.

\bibitem[Zhao et~al.(2020)Zhao, Mopuri, and Bilen]{IDLG}
Bo~Zhao, Konda~Reddy Mopuri, and Hakan Bilen.
\newblock idlg: Improved deep leakage from gradients.
\newblock \emph{ArXiv}, abs/2001.02610, 2020.

\bibitem[Zhou et~al.(2020)Zhou, Chen, Hong, Wu, and Banerjee]{zhou2020private}
Yingxue Zhou, Xiangyi Chen, Mingyi Hong, Zhiwei~Steven Wu, and Arindam
  Banerjee.
\newblock Private stochastic non-convex optimization: Adaptive algorithms and
  tighter generalization bounds.
\newblock \emph{arXiv preprint arXiv:2006.13501}, 2020.

\bibitem[Zhu \& Han(2020)Zhu and Han]{deep_leakage}
Ligeng Zhu and Song Han.
\newblock Deep leakage from gradients.
\newblock In Qiang Yang, Lixin Fan, and Han Yu (eds.), \emph{Federated Learning
  - Privacy and Incentive}, volume 12500 of \emph{Lecture Notes in Computer
  Science}, pp.\  17--31. Springer, 2020.

\bibitem[Zhu et~al.(2021)Zhu, Dong, and Wang]{zhu2021optimal}
Yuqing Zhu, Jinshuo Dong, and Yu-Xiang Wang.
\newblock Optimal accounting of differential privacy via characteristic
  function.
\newblock \emph{arXiv preprint arXiv:2106.08567}, 2021.

\bibitem[Ziller et~al.(2021)Ziller, Usynin, Braren, Makowski, Rueckert, and
  Kaissis]{ziller2021medical}
Alexander Ziller, Dmitrii Usynin, Rickmer Braren, Marcus Makowski, Daniel
  Rueckert, and Georgios Kaissis.
\newblock Medical imaging deep learning with differential privacy.
\newblock \emph{Scientific Reports}, 11\penalty0 (1):\penalty0 1--8, 2021.

\end{thebibliography}
\bibliographystyle{iclr2022_conference}
\newpage
\appendix
\section{Appendix}
\label{apped:theorem1}
\subsection{Unstable Updates}
\label{app:unstable}
\begin{figure}[H]
    \centering
    \includegraphics[width = 0.33\textwidth]{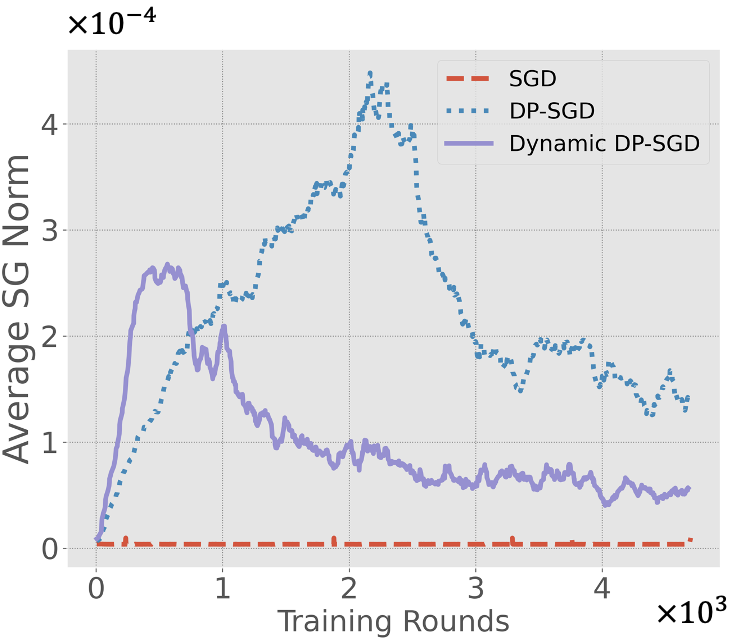}
    \caption{Experiments on Fashion-MNIST for both DP-SGD and the proposed dynamic DP-SGD with the same corresponding  settings as that in Table~\ref{mnist_res}. In comparison to SGD, DP-SGD is unstable, and the stochastic gradient (SG) norm has a ramp-up period.
However, the SG norm is stabilized by the dynamic DP-SGD.
    }
    \label{fig:Gradient_Norm-2}
\end{figure}
\subsection{GDP Preliminary}
\label{app:GDP-Pre}
We first introduce some background about GDP.
Let $P$ and $Q$ denote the distributions of $M(X)$ and $M\left(X^{\prime}\right)$ with $X\sim X^{\prime}$, and let $\phi$ be any (possibly randomized) rejection rule for testing $H_{0}: P$ against $H_{1}: Q$. With these in place, \citet{dong2021gaussian} defines the trade-off function of $P$ and $Q$ as
\begin{equation}
\begin{aligned}
\text{T}(P, Q):[0,1] & \mapsto[0,1] \\
\alpha & \mapsto \inf _{\phi}\left\{1-\mathbb{E}_{Q}[\phi]: \mathbb{E}_{P}[\phi] \leqslant \alpha\right\}.
\end{aligned}
\end{equation}
Above,  $\mathbb{E}_{P}[\phi]$ and  $1-\mathbb{E}_{Q}[\phi]$  are type I and type II errors of the rejection rule  $\phi$, respectively.
It is shown that $\text{T}(P,Q)\geq \text{T}(\mathcal N(0,1),\mathcal N(\mu,1))\triangleq G_{\mu}$, which is referred to as  $\mu$-GDP.

In each step of the  DP-SGD in (\ref{dpsgd}) with the Gaussian mechanism, it achieves $\mu$-GDP with $\mu = \frac{C}{\sigma}$.
Consider the  sampling scheme $\operatorname{PS}(X)$ that each individual data sample $(x, y)$ is subsampled independently with probability $p$ from the training set to construct $X_t$.
It is shown in \citep{bu2019deep} that
given two neighboring datasets $X$ and 
$X^{\prime}$, 
if a randomized mechanism $\mathcal M$ is $G_{\mu}$-{DP},  then
\begin{equation}
\label{GDP-sampleing}
\text{T}\left(\mathcal M \circ \operatorname{PS} (X), \mathcal M \circ \operatorname{PS}\left(X^{\prime}\right)\right) \geqslant p G_{\mu}+(1-p) \mathrm{Id},
\end{equation}
where $\operatorname{Id}(x) = 1 - x$.
Then after a large enough $T$ steps, a Berry-Esseen style CLT result is shown by \cite{bu2019deep} that as $T\to +\infty$ and $p\sqrt{T}\to$ a constant,  the composition of the r.h.s. of (\ref{GDP-sampleing}) converges to a $G_{\mu_{\text{tot}}}$-DP with
\begin{equation}
\label{CLT1}
        \mu_{\text{tot}} = p\sqrt{T(e^{\mu^2}-1)}.
\end{equation}
\subsection{Proof of Theorem \ref{MU-CLT}}
Following the proof of $f$-DP central limit theorem in \citep{bu2019deep}, we have the following  definitions given a function $f$:
\begin{eqnarray}
\mathrm{kl}(f) &\triangleq&
-\int_{0}^{1} \log \left|f^{\prime}(x)\right| \mathrm{d} x, 
\label{def1}
\\ \widetilde{\mathrm{kl}} (f) &\triangleq&
\int_{0}^{1}\left|f^{\prime}(x)\right| \log \left|f^{\prime}(x)\right| \mathrm{d} x, \\ \kappa_{2}(f) &\triangleq&
\int_{0}^{1} \log ^{2}\left|f^{\prime}(x)\right| \mathrm{d} x, \\ \widetilde{\kappa}_{2}(f) &\triangleq
&
\int_{0}^{1}\left|f^{\prime}(x)\right| \log ^{2}\left|f^{\prime}(x)\right| \mathrm{d} x, \\ \kappa_{3}(f) &\triangleq&
\int_{0}^{1}|\log | f^{\prime}(x)||^{3} \mathrm{d} x, \\ 
\widetilde{\kappa}_{3}(f) &\triangleq&
\int_{0}^{1}\left|f^{\prime}(x)\right| \cdot|\log | f^{\prime}(x)||^{3} \mathrm{d} x\label{def6}.
\end{eqnarray}

Let $\left\{f_{n i}: 1 \leqslant i \leqslant n\right\}_{n=1}^{\infty}$ be a triangular array of  trade-off functions  and assume the following limits for some constants $K \geqslant 0$ and $s>0$ as $n \rightarrow \infty$:
\begin{enumerate}
\item
 $\sum_{i=1}^{n} \mathrm{k} \mathrm{l}\left(f_{n i}\right)+\widetilde{\mathrm{k}} \mathrm{l}\left(f_{n i}\right) \rightarrow K$,
\item
 $\max _{1 \leqslant i \leqslant n} \mathrm{kl}\left(f_{n i}\right) \rightarrow 0, \quad \max _{1 \leqslant i \leqslant n} \widetilde{\mathrm{k}} 1\left(f_{n i}\right) \rightarrow 0$,
\item
$\sum_{i=1}^{n} \kappa_{2}\left(f_{n i}\right) \rightarrow s^{2}, \quad \sum_{i=1}^{n} \widetilde{\kappa}_{2}\left(f_{n i}\right) \rightarrow s^{2}$,
\item 
$\sum_{i=1}^{n} \kappa_{3}\left(f_{n i}\right) \rightarrow 0, \quad \sum_{i=1}^{n} \widetilde{\kappa}_{3}\left(f_{n i}\right) \rightarrow 0$,
\end{enumerate}
it is shown in
\citep{bu2019deep} that
\begin{equation}
\label{AsyCLT}
\lim _{n \rightarrow \infty} f_{n 1} \otimes f_{n 2} \otimes \cdots \otimes f_{n n}(\alpha)=G_{K / s}(\alpha)
\end{equation}
uniformly for all $\alpha \in[0,1]$.
Let $g_{t}(x)=-{G}_{\mu_t}^{\prime} -1=\left|{G}_{\mu_t}^{\prime}\right|-1$,
where the second equation is due to the definition of the tradoff function. Then by replacing $n$ by $t$,  equations from (\ref{def1}) to (\ref{def6}) are reformulated as:
\begin{eqnarray} 
\label{kl}
\mathrm{kl}\left(f_{{p,t}}\right)
&=&-\int_{0}^{1} \log (1+p g_{t}(x)) \mathrm{d} x, \\
\label{kltilde}
\widetilde{\mathrm{kl}} \left(f_{p,t}\right) &=&\int_{0}^{1}(1+p g_{t}(x)) \log (1+p g_{t}(x)) \mathrm{d} x, \\
\kappa_{2}\left(f_{p,t}\right) 
&=&\int_{0}^{1}[\log (1+p g_{t}(x))]^{2} \mathrm{d} x, \\
\widetilde{\kappa}_{2}\left(f_{p,t}\right) &=&\int_{0}^{1}(1+p g_{t}(x))[\log (1+p g_{t}(x))]^{2} \mathrm{d} x,
\\
\kappa_{3}\left(f_{p,t}\right)
&=&\int_{0}^{1}[\log (1+p g_{t}(x))]^{3} \mathrm{d} x, \\
\widetilde{\kappa}_{3}\left(f_{p,t}\right) &=&\int_{0}^{1}(1+p g_{t}(x))[\log (1+p g_{t}(x))]^{3} \mathrm{d} x. \end{eqnarray}

According to  (\ref{AsyCLT}), we need to compute 
$\sum_{t=1}^{T}\left(\mathrm{kl}\left(f_{p,t}\right)+\widetilde{\mathrm{k}} \mathrm{l}\left(f_{p,t}\right)\right)$ and
$
\sum_{t=1}^{T} \kappa_{2}\left(f_{p,t}\right)$ to obtain the CLT.
Let $p\sqrt{T}\to v$ and $p\to 0^{+}$, we compute $K$ by
\begin{equation}
\label{K}
\begin{split}
K=&\lim_{
\begin{subarray}{c}
p \rightarrow 0^{+}\\p\sqrt{T}\to v
\end{subarray}
} \sum_{t=1}^{T}\left(\mathrm{kl}\left(f_{p,t}\right)+\widetilde{\mathrm{kl}}\left(f_{p,t}\right)\right)\\
&=\lim _{\begin{subarray}{c}
p \rightarrow 0^{+}\\p\sqrt{T}\to v
\end{subarray}}
\sum_{t=1}^{T}
p\int_{0}^{1} g_{t}(x) \cdot  \log (1+p g_t(x)) \mathrm{d} x \\ 
&=
\sum_{t=1}^{T}
\int_{0}^{1} \frac{v^2}{T} \cdot g_{t}(x) \cdot \lim _{p \rightarrow 0^{+}} \frac{1}{p} \log (1+p g_{t}(x)) \mathrm{d} x \\
&=\frac{v^2}{T}\sum_{t=1}^{T}\int_{0}^{1} g_{t}(x)^{2} \mathrm{d}x\\
&=\frac{v^2}{T}\sum_{t=1}^{T}\left(\text e^{\mu_t^2}-1\right).
\end{split}
\end{equation}
We  further compute
 $s$ following  the same procedure:
\begin{equation}
\begin{split}
\label{s_square}
s^2 =& \sum_{t=1}^{T}
\kappa_{2}\left(f_{p,t}\right)\\
=&
\frac{v^2}{T}
\sum_{t=1}^{T}
\int_{0}^{1}
\lim _{p \to 0^{+}}
\left[\frac{1}{p} \log (1+p g_t(x))\right]^{2} \mathrm{d} x\\
=&\frac{v^2}{T}
\sum_{t=1}^{T}
\int_{0}^{1} g_{t}(x)^{2} \mathrm{d}x\\
=&\frac{v^2}{T}
\sum_{t=1}^{T}\left(\text e^{\mu_t^2}-1\right).
\end{split}
\end{equation}
Substituting (\ref{K}) and (\ref{s_square}) into  (\ref{AsyCLT}), we have
\begin{equation}
\lim _{T \rightarrow \infty} f_{p,t}\otimes_{t=1}^{T}
=G_{\mu}
\end{equation}
with
\begin{equation}
  \mu=\frac{K}{s} 
  =p
  \sqrt{\sum_{t=1}^{T}\left(\text e^{\mu_t^2}-1\right)}.
\end{equation}

\subsection{Dynamic DP-Adam}
\label{app:da-adam}
\begin{algorithm}[H]
%\begin{algorithm}[htbp]
\caption{Dynamic DP-Adam Algorithm}
\label{alg:DP-Adam}
\begin{algorithmic}[1]
\REQUIRE  DP budget $(\epsilon,\delta)$, sampling rate $p$, momentum parameters $(\beta_1,\beta_2)$, a small constant $z>0$, and hyper-parameters:  $\rho_{\mu}$, $\rho_{c}$ and $C_0$.
% \STATE Compute $\mu_{\text{tot}}$ according to (\ref{compute_mu}) \Note{SC: do we need this step?}
\STATE Compute 
$\mu_0 $ in Algorithm~\ref{alg:mu_allocation} 
\FOR{$t=1, \ldots, T$}
%\STATE Compute $C_t$,$\mu_t$ according to \ref{alg:dynamic_noise}
% \STATE Compute $C_t =   (\rho_{\mu_2})^{-\frac{t}{T}}\cdot C_0$.
\STATE Compute $C_t =   (\rho_{c})^{-\frac{t}{T}}\cdot C_0$ in (\ref{sens_decay})
\STATE Calibrate noise :
$\sigma_t = \frac{C_0}{\mu_0}(\rho_{\mu}\cdot\rho_c)^{-\frac{t}{T}}$
% according to 
% (\ref{sigmat_compute_sens_decay}) by replacing  $\rho_c$ with $\rho_{\mu}\cdot\rho_c$. %by substituting $C_t$ into (\ref{noise_power_decay})
% \STATE $\sigma_t = \frac{\rho_{\mu_2}}{\rho_{\mu_2}}\frac{C_0}{\mu_0}$
\STATE 
Sample $ X_t\in X$ with sampling rate $p$ and sample noise $\xi_{t}\sim \mathcal N(0,\sigma^2_{t}I)$. 
\STATE Compute:
    $\widetilde g_t=\frac{1 }{\left|X_t\right|}\big[\xi_{t}+$
    $ 
    \sum_{x\in X_{ t}}\text{CL}\left(g_x;C_t\right)
    \big]$ with $\text{CL}(\cdot)$ in (\ref{clipping}) 
\STATE Compute biased first  momentum:
$m_{t} = \beta_{1} m_{t-1}+\left(1-\beta_{1}\right) \tilde{g}_{t}$
\STATE Compute biased second  momentum:
$u_{t} \leftarrow \beta_{2} u_{t-1}+\left(1-\beta_{2}\right)\left(\tilde{g}_{t} \odot \tilde{g}_{t}\right)$
\STATE
Compute bias-corrected first moment:
$\widehat{m}_{t} = m_{t} /\left(1-\beta_{1}^{t}\right)$ 
\STATE 
Compute bias-corrected second  moment 
$\widehat{v}_{t} = v_{t} /\left(1-\beta_{2}^{t}\right)$ 
\STATE Compute:
$w_{t} =\widehat m_{t} /\left(\sqrt{\widehat u_{t}}+z\right)$
% \STATE
%  $\theta_t^{(\ell)} =  \theta_{t-1}^{(\ell)} - \eta 
%     \frac{1}{\left|X_t\right|} \left(\widetilde g_t + \xi_{t}\right)$ with 
%     $\xi_{i,t}\sim \mathcal N(0,\sigma^2_{i,t})$
\STATE Compute:
    $\theta_{t+1} =  \theta_{t} - \eta w_t$
\ENDFOR
\end{algorithmic}
\end{algorithm}

\subsection{Proof of Corollary \ref{thm: dp_alg3}}
\label{app:proof-corrollary1}

From the definition Gaussian DP in Eq.(~\ref{mu-def}), we know each step in Algorithm 3 is $G_{\mu_t}$-DP and 
\begin{align}
    \mu_t = \frac{C_t}{\widetilde\sigma_t\widetilde\sigma }.
\end{align}

Further, by Theorem \ref{MU-CLT}, we know overall Algorithm 3 is $G_{\mu}$-DP with
\begin{align}
    \mu = p\sqrt{\sum_{t=1}^T (e^{\mu_t^2} -1) }.
\end{align}

Because $e^x < 1 + 2x$ for $0 \leq x\leq 1$,  we know
\begin{align}
    \mu < p\sqrt{2\sum_{t=1}^T \mu_t^2,}\quad \text{when $\mu_t^2 \leq 1$} . 
\end{align}

Given a total privacy budget $\mu_{\text{tot}}$ that bounds the above $\mu$, we have 

\begin{align}
    \mu < p\sqrt{2\sum_{t=1}^T \mu_t^2}  = p\sqrt{2\sum_{t=1}^T \frac{C_t^2}{\widetilde\sigma^2\widetilde\sigma_t^2}}   \leq \mu_{\text{tot}}.
\end{align}

Rearranging the last inequality, we get
\begin{align}
    \widetilde\sigma \geq  p \frac{1}{\mu_{\text{tot}}}\sqrt{2\sum_{t=1}^T \frac{C_t^2}{\widetilde\sigma_t^2}},
\end{align}
which is  sufficient to guarantee Algorithm 3 to satisfy $G_{\mu_{\text{tot}}}$-DP.

\subsection{Proof of Theorem \ref{thm: utl_alg3}}
\label{appendix:thm-utl_alg3}
By the update rule in Algorithm 3, we know
\begin{align}
    \theta_{t=1} = \theta_{t} - \frac{\eta}{pN} \widetilde\sigma_t \xi_t - \eta g_{t}.
\end{align}

By the assumption that the objective function $L(\cdot)$ is $Q$-smooth, we have
\begin{align}\label{eq: smooth}
    \mathbb  E_{t}[L(\theta_{t+1})] \leq &  L(\theta_{t})  + \mathbb E_{t}[ \langle \nabla L(\theta_{t}) , (\theta_{t+1} - \theta_{t})\rangle] \nonumber \\
    &+ \mathbb E_{t}[ \frac{Q}{2}\|\theta_{t+1} -  \theta_{t}\|^2] \nonumber \\
    \leq &  L(\theta_{t})  -\eta \langle \nabla L(\theta_{t}), \mathbb E_{t}[g_{t}] \rangle + \frac{Q}{2} \eta^2 \mathbb E_{t}[ \| g_{t}\|^2 ] \nonumber \\
    &+ \frac{Q}{2} \frac{\eta^2}{N^2p^2}\widetilde\sigma^2\widetilde\sigma_{t}^2, 
\end{align}
where $\mathbb E_t$ denotes expectation taken over randomness at step $t$, including gradient sampling and noise sampling.
Taking the sum of $t$ from $1$ to $T$ and the expectation of all steps, we get 
\begin{align}
    & \mathbb  E[L(\theta_{T+1})]  - \mathbb E[L(\theta_{1})] \nonumber \\
    \leq &    -\eta \mathbb E\left [\sum_{t=1}^T \langle \nabla L(\theta_{t}), \mathbb E_{t}[g_{t}] \rangle \right] + \frac{Q}{2} \eta^2 \sum_{t=1}^T \mathbb E[ \| g_{t}\|^2 ] \nonumber \\
    &+ \sum_{t=1}^T \frac{Q}{2} \frac{\eta^2}{N^2p^2}\widetilde\sigma^2\widetilde\sigma_{t}^2. 
\end{align}

Rearranging and dividing both sides by $T\eta$, we have

\begin{align}\label{eq: after_tele}
    & \frac{1}{T} \mathbb E\left [\sum_{t=1}^T \langle \nabla L(\theta_{t}), \mathbb E_{t}[g_{t}] \rangle \right] \nonumber \\
    \leq &  \frac{1}{\eta T} (\mathbb E[L(\theta_{1})]  - \mathbb  E[L(\theta_{T+1})])    + \frac{1}{T}\frac{Q}{2} \eta \sum_{t=1}^T \mathbb E[ \| g_{t}\|^2 ] \nonumber \\
    &+ \frac{1}{T}\sum_{t=1}^T \frac{Q}{2} \frac{\eta}{N^2p^2}\widetilde\sigma^2\widetilde\sigma_{t}^2. 
\end{align}

Further, we know
\begin{align}
    \mathbb E_{t}[g_{t}] = & \mathbb E[ \frac{1}{p\left|X\right|}\big[
    \sum_{x\in X_{ t}}\text{CL}\left(g_x;C_t\right)
    \big]] \nonumber \\
    = & \mathbb E_{X_t}[ \frac{1}{p\left|X\right|}\big[
    \sum_{x\in X_{ t}}\text{CL}\left(g_x;C_t\right)
    \big]] \nonumber \\
    \overset{(i)}{=} & \frac{1}{p\left|X\right|} p\left|X\right| \mathbb E_{x}[ \text{CL}\left(g_x;C_t\right)
    ] \nonumber \\
    \overset{(ii)}{=} &  \sum_{i=1}^N\frac{1}{N} \text{CL}\left(g_{x_i};C_t\right) \nonumber \\
    \overset{(iii)}{=} &  \sum_{i=1}^N\frac{1}{N} \mathbbm{1} \left[\|g_{x_i}\|\leq C_t\right]g_{x_i} \nonumber \\
    & +  \sum_{i=1}^N\frac{1}{N} \mathbbm{1} \left[\|g_{x_i}\|> C_t\right] g_{x_i}\frac{C_t}{\|g_{x_i}\|} \nonumber \\
    {=} &  \sum_{i=1}^N\frac{1}{N} g_{x_i} \nonumber \\
    & +  \sum_{i=1}^N\frac{1}{N} \mathbbm{1} \left[\|g_{x_i}\|> C_t\right] g_{x_i}\left(\frac{C_t}{\|g_{x_i}\|}-1\right), \nonumber \\
\end{align}
where $(i)$ is due to each element in $X$ has $p$ probability being in $X_{t}$ and $\mathbb E_x$ is taken over a uniform distribution of $x \in X$; $(ii)$ is because we assumed that there is  $N$ samples of $x$; $(iii)$ is due to the definition of clipping operation.

Recall the fact that $\nabla L(\theta_t) = \frac{1}{N}\sum_{i=1}^N g_{x_i}$ and the assumption $\|g_{x_i}\| \leq G$, we have 
\begin{align}\label{eq: bias_prob}
    &\|\mathbb E_{t}[g_{t}] - \nabla L(\theta_t)\| \nonumber\\
    = & \left\|\sum_{i=1}^N\frac{1}{N} \mathbbm{1} \left[\|g_{x_i}\|> C_t\right] g_{x_i}\left(\frac{C_t}{\|g_{x_i}\|}-1\right)\right\|\nonumber \\
    \leq & \sum_{i=1}^N\frac{1}{N} \mathbbm{1} \left[\|g_{x_i}\|> C_t\right] G \nonumber \\
    = & P_t(C_t) G, 
\end{align}
where $P_t(C_t)$ denotes the probability of a per-sample gradient being clipped at step $t$ given $\theta_t$ and $X$.

Then we know
\begin{align}\label{eq: split_bias}
    &\frac{1}{T} \mathbb E\left [\sum_{t=1}^T \langle \nabla L(\theta_{t}), \mathbb E_{t}[g_{t}] \rangle \right] \nonumber \\
    = & \frac{1}{T} \mathbb E\left [\sum_{t=1}^T \langle \nabla L(\theta_{t}), \nabla L(\theta_{t}) - \nabla L(\theta_{t}) + \mathbb E_{t}[g_{t}] \rangle \right] \nonumber \\
    \geq &  \frac{1}{T} \mathbb E\left [\sum_{t=1}^T \| \nabla L(\theta_{t})\|^2 \right] \nonumber \\
    & - \frac{1}{T} \mathbb E\left [\sum_{t=1}^T \| \nabla L(\theta_{t})\|\| \nabla L(\theta_{t}) - \mathbb E_{t}[g_{t}] \| \right] \nonumber \\
    \geq & \frac{1}{T} \mathbb E\left [\sum_{t=1}^T \| \nabla L(\theta_{t})\|^2 \right] \nonumber \\
    & - \frac{1}{T} \mathbb E\left [\sum_{t=1}^T \| \nabla L(\theta_{t})\|P_t(C_t) G \right].
\end{align}

Substituting \eqref{eq: split_bias} into the LHS of \eqref{eq: after_tele}, we get

\begin{align}\label{eq: final}
    & \frac{1}{T} \mathbb E\left [\sum_{t=1}^T \| \nabla L(\theta_{t})\|^2 \right] \nonumber \\
    \leq &  \frac{1}{\eta T} (\mathbb E[L(\theta_{1})]  - \mathbb  E[L(\theta_{T+1})])    + \frac{1}{T}\frac{Q}{2} \eta \sum_{t=1}^T \mathbb E[ \| g_{t}\|^2 ] \nonumber \\
    &+ \frac{1}{T}\sum_{t=1}^T \frac{Q}{2} \frac{\eta}{N^2p^2}\widetilde\sigma^2\widetilde\sigma_{t}^2  + \frac{1}{T} \mathbb E\left [\sum_{t=1}^T \| \nabla L(\theta_{t})\|P_t(C_t) G \right],
\end{align}
which finish the proof for the first claim.

By substituting $\widetilde\sigma = p \frac{1}{\mu_{\text{tot}}}\sqrt{2\sum_{t=1}^T \frac{C_t^2}{\widetilde\sigma_t^2}}$ from Corollary \ref{thm: dp_alg3}, assume $\widetilde\sigma_t = \Theta(1)$ and $L$ is lower bounded, we can further get
\begin{align}
    & \frac{1}{T} \mathbb E\left [\sum_{t=1}^T \| \nabla L(\theta_{t})\|^2 \right] \nonumber \\
    \leq &  \frac{1}{\eta T} (\mathbb E[L(\theta_{1})]  - \mathbb  E[L(\theta_{T+1})])    + \frac{1}{T}\frac{Q}{2} \eta \sum_{t=1}^T \mathbb E[ \| g_{t}\|^2 ] \nonumber \\
    &+ {Q} \frac{\eta}{N^2\mu_{\text{tot}}^2} \frac{1}{T}\sum_{t=1}^T \widetilde\sigma_t^2  {\sum_{t=1}^T \frac{C_t^2}{\widetilde\sigma_t^2}} \nonumber \\
    & + \frac{1}{T} \mathbb E\left [\sum_{t=1}^T \| \nabla L(\theta_{t})\|P_t(C_t) G \right] \nonumber \\
    = & \mathcal O\left(\frac{1}{\eta T}\right)    + \mathcal O(\eta) +\mathcal O \left( \frac{\eta T}{N^2\mu_{\text{tot}}^2} \right) \nonumber \\
    & + \underbrace{\frac{1}{T} \mathbb E\left [\sum_{t=1}^T \| \nabla L(\theta_{t})\|P_t(C_t) G \right]}_{\text{bias term}}
\end{align}
and after setting $\eta T = N \mu_{\text{tot}}$ and $\eta = \frac{1}{N\mu_{\text{tot}}}$, we can reach the conclusion that the above bound is $O(\frac{1}{N \mu_{\text{tot}}}) + \text{bias term}$ which finishes the whole proof.

\subsection{On the optimal solution of minimizing Eq.~\eqref{eq:dp_variance-D1}}\label{app: opt_sol}
The minimization problem is rewritten below ignoring the irrelevant constants
\begin{align}\label{eq: dp_variance} 
   \min_{\widetilde\sigma_1,...,\widetilde\sigma_T} \sum_{t=1}^T \widetilde\sigma_t^2 {\sum_{t=1}^T \frac{C_t^2}{\widetilde\sigma_t^2}}.
\end{align}

The above problem can be transformed into an equivalent problem raised in Theorem 4.2 and Remark 4.1 of \citet{wu2021adaptive} which we restate below
\begin{align}\label{eq: prob_refer}
    \min_{a_1,...,a_T}  \sum_{t=1}^T \frac{a_t^2}{b_t^2} \sum_{t=1}^T \frac{1}{a_t^2}
\end{align}

And according to Remark 4.1 in \citet{wu2021adaptive}, an optimal solution to \eqref{eq: prob_refer} is $a_t^2 = b_t$ which yields an optimal value of $( \sum_{t=1}^T\frac{1}{b_t})^2$.
Going back to \eqref{eq: dp_variance}, we can define $a_t = \sigma_t/C_t$ and $b_t=1/C_t$ to transform \eqref{eq: dp_variance} into \eqref{eq: prob_refer}. Thus, we know an optimal solution to \eqref{eq: dp_variance} would be 
\begin{align}
    a_t^2 = \frac{\widetilde\sigma_t^2}{C_t^2} =  b_t = \frac{1}{C_t},
\end{align}
which gives
\begin{align}
   \widetilde \sigma_t =  \sqrt{C_t}.
\end{align}

\subsection{Extension of Theorem \ref{thm: utl_alg3} with local curvature assumption}\label{app: local_cur}

In this section, we present a different version of Theorem \ref{thm: utl_alg3} with a local curvature assumption
\begin{align}\label{eq: local_curv}
    & \mathbb E_t[L(\theta_{t+1})] - L(\theta)  \nonumber \\
    \leq & \langle \nabla L(\theta_t), \mathbb E_t[\theta_{t+1}-\theta_t] \rangle + \frac{Q_t}{2} \mathbb E[\|\theta_{t+1}-\theta_t\|^2].
\end{align}
This assumption is not standard in optimization but the intuition is quite straightforward. The assumption can be understood as the neighborhood region around $\theta_t$ has a curvature bounded by $Q_t$, and the diameter of the   neighborhood is vaguely defined by some statistics about $\theta_{t+1}-\theta_t$, which in turn depends on the algorithm choice. Overall, this is an optimistic assumption on both the algorithm and the problem. The motivation of such an assumption is to take account for the fact that the local curvatures of different iterates in neural network training are typically quite different.

With this assumption, we can  rework Theorem \ref{thm: utl_alg3} to obtain  Theorem \ref{thm: local_curv}.

\begin{theorem}\label{thm: local_curv}
Denote the objective function as $ L(\theta) = \frac{1}{N}\sum_{i=1}^N f(\theta, x_i)$ and the clipped stochastic gradient at step $t$ as $g_t = \frac{1}{p\left|X\right|}\big[
    \sum_{x\in X_{ t}}\text{CL}\left(g_x;C_t\right)
    \big]$ where $g_x = \frac{\partial f(\theta_t, x) }{\partial \theta_t} $.  
    
    Assume \eqref{eq: local_curv} holds and each per-sample gradient is upper-bounded, i.e. $\|g_x\| \leq G, \forall t$.  General dynamic DP-SGD (Algorithm 3) 
satisfies the following utility guarantee
{\small
\begin{align}\label{eq: sgd_bound_local}
      & \mathbb E\left[ \frac{1}{T} \sum_{t=1}^T \| \nabla   L(\theta_{t})\|^2 \right] \nonumber \\
      & \leq   \frac{1}{\eta T}  \mathbb  E[L(\theta_{1})] - L(\theta_{T+1})] 
     +    \frac{\eta}{T}  \sum_{t=1}^T  \frac{Q_t}{2} \mathbb E[ \| g_{t}\|^2]  \nonumber \\ 
      & + \underbrace{\mathbb E \left[\frac{1}{T} \sum_{t=1}^T \| \nabla   L(\theta_{t})\| P_t(C_t) G \right]}_{\text{bias term}}+  \frac{\eta}{p^2N^2}\underbrace{ \frac{1}{T }\sum_{t=1}^T \frac{Q_t}{2}  \widetilde\sigma^2\widetilde\sigma_{t}^2  }_{D_2}.
\end{align}
}%
where $P_t(C_t)$ is the probability of a per-sample gradient being clipped with threshold $C_t$ at $\theta_t$, and the expectations are take over all randomness including gradient sampling and noise sampling. 
\end{theorem}

\textbf{Proof:} The proof follows the same procedure as that of Theorem \ref{thm: utl_alg3} with $Q$ in \eqref{eq: smooth} replaced by $Q_t$, i.e., we get the following version of \eqref{eq: smooth}
\begin{align}\label{eq: smooth_local}
    \mathbb  E_{t}[L(\theta_{t+1})] \leq &  L(\theta_{t})  + \mathbb E_{t}[ \langle \nabla L(\theta_{t}) , (\theta_{t+1} - \theta_{t})\rangle] \nonumber \\
    &+ \mathbb E_{t}[ \frac{Q_t}{2}\|\theta_{t+1} -  \theta_{t}\|^2] \nonumber \\
    \leq &  L(\theta_{t})  -\eta \langle \nabla L(\theta_{t}), \mathbb E_{t}[g_{t}] \rangle + \frac{Q_t}{2} \eta^2 \mathbb E_{t}[ \| g_{t}\|^2 ] \nonumber \\
    &+ \frac{Q_t}{2} \frac{\eta^2}{N^2p^2}\widetilde\sigma^2\widetilde\sigma_{t}^2. 
\end{align}
All the remaining steps are the same as those in Theorem \ref{thm: utl_alg3} with $Q$ replaced and thus we omit them.

\textbf{Remark on the variance term $D_2$:} It is straightforward to see that just like in Theorem \ref{thm: utl_alg3}, we can minimize the corresponding variance term $D_2$ by substituting the $\widetilde\sigma$ expression from Theorem \ref{thm: dp_alg3} and tuning $\widetilde\sigma_1,...,\widetilde\sigma_T$. However, the minimization problem is slightly different due to $Q_t$, the problem is written below ignoring irrelevant constants
\begin{align}
   \min_{\widetilde\sigma_1,...,\widetilde\sigma_T}  \sum_{t=1}^T {Q_t}  \widetilde\sigma_t^2{\sum_{t=1}^T \frac{C_t^2}{\widetilde\sigma_t^2}}.
\end{align}

This problem can also be transformed in to the form of \ref{eq: prob_refer} by defining  $a_t = \sigma_t/C_t$ and $b_t=1/(C_t\sqrt{Q_t})$. And according to the optimal solution $a_t^2 = b_t$, we have
\begin{align}
    a_t^2 = \frac{\sigma_t^2}{C_t^2} =  b_t = \frac{1}{C_t\sqrt{Q_t}},
\end{align}
which indicates
\begin{align}
    \widetilde\sigma_t =  \frac{\sqrt{C_t}}{{Q_t^{1/4}}},
\end{align}
and thus 
\begin{align}
    \mu_t = \frac{C_t}{\widetilde\sigma \cdot \widetilde\sigma_{t}} \propto \sqrt{C_t} Q_t^{1/4}.
\end{align}

Such a result indicates one should also change privacy budget of different steps according to local curvatures. And if assuming the local curvatures of different steps follow some general rules or distributions, one could design corresponding strategies to allocate the privacy budget. Yet, identifying such rules and distributions is a difficult task and could depend on the application domains. We hope the theoretical analysis here could help inspire future exploration in this direction.

\subsection{Experimental Settings on Different Data Sets}
\label{apped:dataset}
\textbf{MNIST} \citep{mnist}  that contains 60000 training samples and 10,000 testing samples with ten balanced classes. Each gray-scale sample is stored as a $28\times 28$ matrix. The classification model used in our experiments consists of two convolution layers and two fully connected layers. Following each convolution layer is a max-pooling layer with a pooling size of $2\times2$. For activations, we use ReLU and for classification, we use softmax. Cross-entropy loss is used.

In the experiment, initial clipping value is set to 1.5 and learning rate is set to 0.15. We set $p = \frac{250}{60000}$ for the subsampling with independent Bernoulli trial. 
% When training with poisson sampling instead of shuffling, the term epoch is no longer valid as visit to each sample is not guaranteed. We want to simulate to train 20 epochs so we compute with poisson sampling rate to get that we are supposed to train 4800 rounds. 
All the results are the average over 5 times repeated experiments.

\textbf{FashionMNIST} is a dataset of Zalando's article images, serving as an alternative to MNIST dataset for benchmarking machine learning algorithms. It shares the same image size and structure of training and testing splits. As a result, we use the same experimental setup as MNIST. The only difference is that we  specified a clipping value of 4 as the initial value.

% \textbf{CIFAR-10} 
% is a dataset consists of 50000 training images and 10000 test images in 10 classes. Each sample is formatted as a 32x32 images in RGB.  We conduct experiments on CIFAR10 with ResNet-18. We set initial clipping value to 15, while remaining other experimental setting the same.
\textbf{IMDB} For natural language processing or text analytics, the IMDB dataset contains 50K movie reviews. This is a binary sentiment classification dataset. We use a three-layer network with one embedding layer and two fully connected layers; this can be viewed as an MLP model because the embedding layer is a special implementation of fully connected layers. We train for 25 epochs with the DP-Adam optimizer, with the initial clipping value set to be 2.

% Please keep in mind that the DP-SGD and proposed dynamic DP-SGD can also be used to obtain the DP-Adam counterparts of Adam due to DP's postprocessing properties.

\textbf{NAME} is a name classification dataset containing person names from 18 countries. It is available on Pytorch NLP tutorials\footnote{https://pytorch.org/tutorials/intermediate/char\_rnn\_classification\_tutorial.html}. 
We train a LSTM model to determine which country the given name belongs to. The name is treated as a sequence, and the characters are fed into LSTM one by one. We use a one-layer LSTM with hidden size 128 and embedding size 64 followed by a fully connected layer, an SGD optimizer with learning rate 2 and we train 50 epochs for each experiment, and the initial clipping value is set to be 1.5. 
% \textbf{SST-2} is The Stanford Sentiment Treebank dataset, it consists of sentences from movie reviews and human annotations of their sentiment. The task is to predict the sentiment of a given sentence. We use the two-way (positive/negative) class split, and use only sentence-level labels.

\textbf{InfiniteMNIST and Federated Learning } InfiniteMNIST is a dataset consisting of massive training samples derived from origin MNIST by applying different types of transformations. Such large scale dataset is suitable for a federated setting. We extract 250K and 500K images as two training data sets and simulate identical number of clients. Thus each client is not enough to train the model, but they can cooperatively learn a model via federated learning.  We use the same network structure as in the experiment for MNIST for each client.

\textbf{Federated Learning Algorithm} 
Fed-SGD algorithm is used for optimization. For each round, we randomly sample clients with sampling rate $p=1\times10^{-3}$ for the case of MNIST-250K and $p=5\times10^{-4}$ for the case of MNIST-500K.  Gradients are computed on each selected client and then sent to the server for aggregation. In this setting, local DP mechanism is required to protect client side gradients, and each client apply clipping and additive noise on local gradients before transmission. The server will aggregate noised gradients, which achieves central DP.
The details is provided in the following  Algorithm~\ref{alg:fed}.  

\begin{algorithm}[h]
\caption{Federated Dynamic DP-SGD Algorithm}
\label{alg:fed}
\begin{algorithmic}[1]
\REQUIRE  Clients set $\mathcal S$, DP budget $(\epsilon,\delta)$, client sampling rate $p$, dataset $X$ = ($X_1,X_2...X_{|\mathcal{S}|}$), training rounds $T$, hyper-parameters:  $\rho_{\mu}$, $\rho_{c}$ and $C_0$.
\STATE Compute 
$\mu_0 $ in Algorithm~\ref{alg:mu_allocation} 
\FOR{$t=0, \ldots, T-1$}
\STATE Compute $C_t =   (\rho_{c})^{-\frac{t}{T}}\cdot C_0$ according to (\ref{sens_decay})
\STATE Calibrate noise :
$\sigma_t = \frac{C_0}{\mu_0}(\rho_{\mu}\cdot\rho_c)^{-\frac{t}{T}}$

% \STATE Compute $(\mu_t,C_t,\sigma_t)$ according to Algorithm (\ref{alg:dynamic_noise})
\STATE 
Sample clients for the $t$-th iteration, i.e., $\mathcal S_t\in \mathcal S$ with Poisson sampling rate $p$.
\FOR{$k \in S_t$}
% \FOR{$x \in X_t$}

\STATE $g_{k,t} = 
    \frac{1}{\left|\mathcal S_t\right|} \left( 
    \sum_{x\in \mathcal S_{t}}\left(g_x;C_t\right)
    + \xi_{k,t}\right)$  with 
    $\xi_{k,t}\sim \mathcal N(0,\sigma^2I)$
% \ENDFOR
\STATE Send $g_{k,t} $ to the server.
\ENDFOR
\STATE    
    Server computes: $\theta_{t+1} =  \theta_{t} - \eta 
    \frac{1}{\left|\mathcal{S}_t\right|} \left( 
    \sum_{k\in \mathcal S_{ t}} g_k\right)$
\ENDFOR

\end{algorithmic}
\end{algorithm}

\subsection{VGG network and Chest Radiographs}
We evaluated VGG on chest radiographs classification task
from the Paediatric Pneumonia dataset  described in~\citet{ziller2021medical}, which  is a binary classification
task, and  the model attempts to predict whether the radiograph shows signs of pneumonia or not.
For the classification task, following that in~\citet{ziller2021medical}, we utilized the same model architecture in the private and nonprivate setting, namely a VGG-11 architecture. Following~\citet{ziller2021medical},  batch normalization layers was disabled for both non-private and DP training.

\end{document}